\documentclass{article}

\usepackage{arxiv}

\usepackage[utf8]{inputenc}
\usepackage[T1]{fontenc}
\usepackage{hyperref}
\usepackage{url}
\usepackage{booktabs}
\usepackage{amsfonts}
\usepackage{nicefrac}
\usepackage[expansion=false]{microtype}
\usepackage{graphicx}
\usepackage{float}
\usepackage{amsmath,amssymb,amsthm}
\usepackage{multirow}
\usepackage{array}
\usepackage{xcolor}
\usepackage{caption}
\usepackage{enumitem}
\usepackage{subcaption}
\usepackage{cleveref}
\usepackage{setspace}
\usepackage{natbib}
\usepackage{doi}
\usepackage{placeins}
\usepackage{adjustbox}
\usepackage{pifont}
\usepackage{tabularx}
\usepackage{makecell}
\usepackage{ragged2e}
\usepackage{threeparttable}
\usepackage{xspace}
\usepackage{mathrsfs}
\usepackage{bm}
\usepackage{pdflscape}
\usepackage{rotating}

\usepackage{algorithm}
\usepackage{algpseudocode}

\usepackage{siunitx}
\newcommand{\cmark}{\ding{51}}
\newcommand{\xmark}{\ding{55}}

\graphicspath{{figures/}}

\title{Language-Guided Hierarchical Locomotion for Small-Scale Bipedal Robots: From Reinforcement Learning to Voice-Controlled Maze Navigation}

\author{
  Rajmeet~Singh$^{*}$ \\
  Mohamed bin Zayed University of Artificial Intelligence\\
  Abu Dhabi, UAE \\
  \texttt{rajmeet.bhourji@mbzuai.ac.ae}
  \And
  Bater~Makhabel \\
  Mohamed bin Zayed University of Artificial Intelligence\\
  Abu Dhabi, UAE \\
  \texttt{Bater.Makhabel@mbzuai.ac.ae}
  \And
  \vspace{-1em}
  \normalsize\textit{$^{*}$ Corresponding author.}
}

\hypersetup{
  pdftitle={Language-Guided Hierarchical Locomotion for Small-Scale Bipedal Robots: From Reinforcement Learning to Voice-Controlled Maze Navigation},
  pdfsubject={Reinforcement Learning, Bipedal Locomotion, Language-Guided Robotics, Small-Scale Robots},
  pdfauthor={Rajmeet Singh, Bater Makhabel},
  pdfkeywords={Bipedal locomotion, Reinforcement learning, Language-guided robotics, Maze navigation, Small-scale robots, Hierarchical control, Voice control},
  colorlinks=true,
  citecolor=blue,
  linkcolor=blue,
  urlcolor=blue
}

\begin{document}
\maketitle

\begin{abstract}
This paper presents a hierarchical control framework for language-guided locomotion of a small-scale bipedal robot. We address three interrelated challenges on the MicroDuck platform, a \ensuremath{25\,\mathrm{cm}}, \ensuremath{800\,\mathrm{g}}, 14-DOF biped actuated by Dynamixel XL330 servos with \ensuremath{0.60\,\mathrm{N\,m}} stall torque: (1)~training stable walking policies under severe torque constraints using Proximal Policy Optimization~(PPO) with terrain curriculum and domain randomization in MuJoCo Warp, (2)~autonomous maze navigation through a decoupled two-layer architecture combining an RL-trained navigator with the walking policy, and (3)~natural language and voice control via a local Large Language Model~(LLM) that translates human commands into velocity primitives. The walking policy achieves a mean reward of 6.96 and generalizes to slopes up to \ensuremath{10^{\circ}}. Systematic ablation studies show that removing terrain curriculum degrades reward by 34\% and removing domain randomization by 22\%. The trained navigation policy achieves 100\% maze-solving success rate (7/7 runs, mean \ensuremath{43.3\,\mathrm{s}}). The LLM interface maps diverse natural language commands including voice input to stable bipedal locomotion with 88\% parse accuracy and \ensuremath{2.6\,\mathrm{s}} end-to-end latency. We further document the negative result that XL330 servos are insufficient for vertical jumping. All experiments use the Battery Actuator Model~(BAM) for high-fidelity servo dynamics. To our knowledge, this is the first demonstration of voice-controlled, LLM-guided hierarchical locomotion on a sub-\ensuremath{1\,\mathrm{kg}} bipedal robot.
\end{abstract}


\section{Introduction}\label{sec:intro}

Bipedal locomotion has advanced substantially through reinforcement learning~(RL), with policies trained in simulation and deployed on platforms ranging from quadrupeds~\citep{lee2020learning,miki2022learning,kumar2021rma} to full-size humanoids~\citep{radosavovic2024real,he2024omnih2o}. These platforms share generous actuator torque margins typically exceeding \ensuremath{10\,\mathrm{N\,m}} per joint for robots weighing \ensuremath{20\,\mathrm{kg}}--\ensuremath{80\,\mathrm{kg}} enabling aggressive locomotion strategies infeasible for small-scale robots near their physical limits~\citep{margolis2023walk}.

The emergence of affordable, open-source bipedal platforms such as the MicroDuck (Pollen Robotics) creates an opportunity to study locomotion under fundamentally different constraints. At \ensuremath{25\,\mathrm{cm}} and \ensuremath{800\,\mathrm{g}}, the MicroDuck is actuated by 14~Dynamixel XL330-M288-T servos producing only \ensuremath{0.60\,\mathrm{N\,m}} stall torque. Walking on flat ground demands careful reward engineering~\citep{bhourji2024rl,marum2024revisiting}; navigating slopes and mazes pushes the platform to its physical envelope; and jumping exceeds it entirely.

Simultaneously, the integration of LLMs with robotic systems has gained attention through SayCan~\citep{ahn2022saycan}, Code as Policies~\citep{liang2023code}, and RT-2~\citep{brohan2023rt2}. For legged robots, Ouyang et al.~\citep{ouyang2024llm} used LLMs for quadruped task planning, and Yao et al.~\citep{yao2024anybipe} proposed AnyBipe for LLM-guided bipedal training. However, the direct grounding of natural language in bipedal locomotion primitives where the LLM translates ``walk forward slowly'' into a velocity command a torque-limited biped can execute remains unexplored.

\begin{figure}[!ht]
\centering
\includegraphics[width=\linewidth]{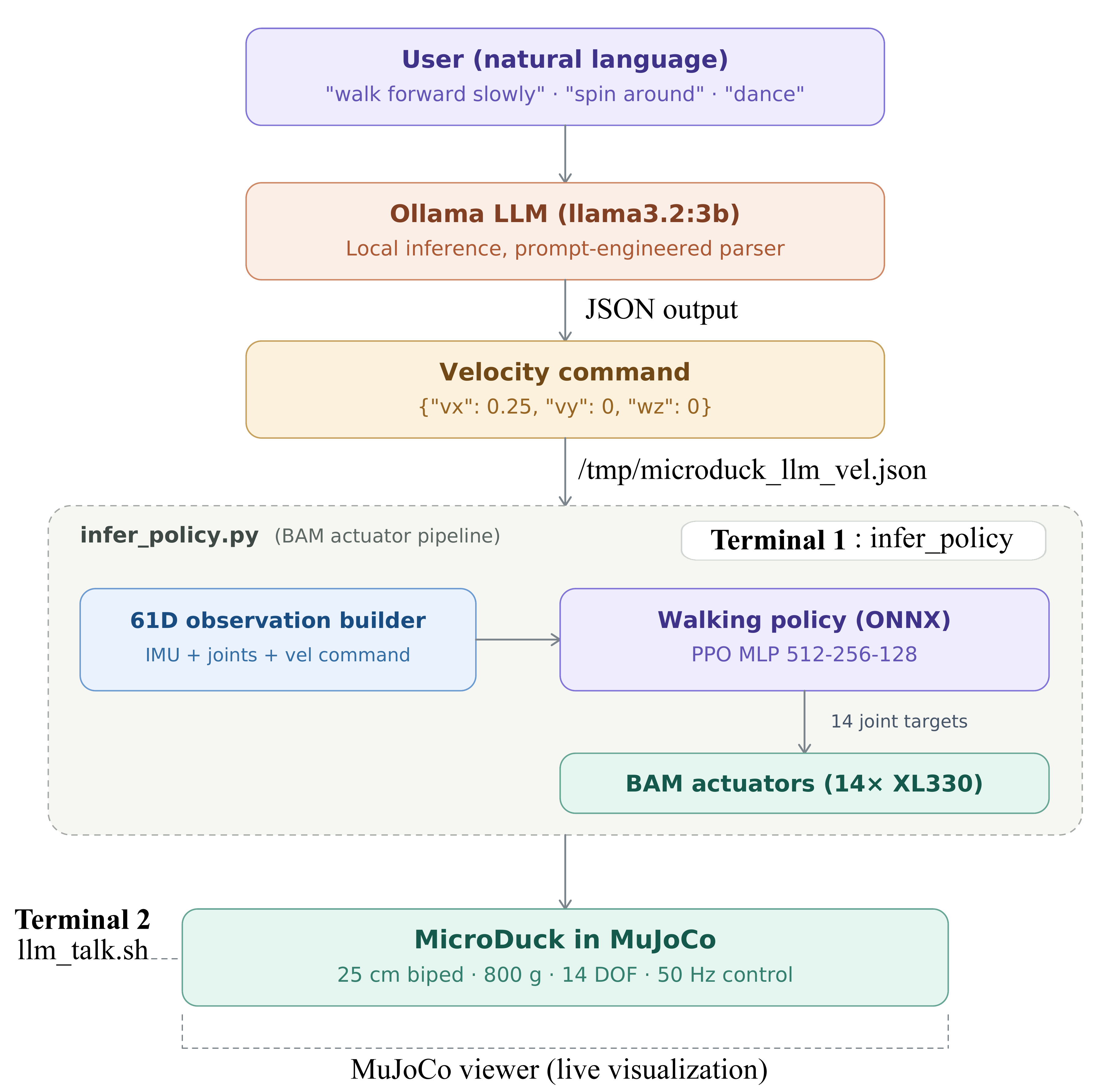}
\caption{Three-layer hierarchical architecture. Language layer (LLM/voice) $\rightarrow$ navigation layer (RL navigator) $\rightarrow$ locomotion layer (walking policy + BAM actuators). Each layer is trained independently; composition occurs through shared velocity commands.}
\label{fig:system}
\end{figure}

This paper addresses this gap through a hierarchical framework (Fig.~\ref{fig:system}) decomposing the problem into three independently trainable layers:
\begin{enumerate}[leftmargin=*]
\item \textbf{Locomotion layer:} A PPO-trained walking policy (61D observations $\rightarrow$ 14~joint targets) with terrain curriculum and domain randomization.
\item \textbf{Navigation layer:} An RL-trained maze navigator (11D observations $\rightarrow$ 2D velocity commands) using rangefinder-based perception.
\item \textbf{Language layer:} A local LLM (Ollama, llama3.2:3b) parsing text or voice commands into velocity primitives.
\end{enumerate}

\subsection{Contributions}
\begin{itemize}[leftmargin=*]
\item {A terrain curriculum and reward architecture for torque-limited bipeds, with ablation quantifying each component (Section~\ref{sec:loco})}.
\item{A decoupled navigation-locomotion architecture achieving 100\% maze-solving rate}.
\item{ The first LLM-guided and voice-controlled bipedal locomotion on a sub-\ensuremath{1\,\mathrm{kg}} platform (Section~\ref{sec:language}).}
\item{ Physical envelope characterization including negative jump results (Section~\ref{sec:results}).}
\item{ An open-source simulation benchmark using MjLab, BAM, and ONNX.}
\end{itemize}

\section{Related Work}\label{sec:related}

\subsection{Reinforcement Learning for Legged Locomotion}
Lee et al.~\citep{lee2020learning} demonstrated robust quadruped locomotion on ANYmal. Miki et al.~\citep{miki2022learning} extended this with perceptive locomotion. Kumar et al.~\citep{kumar2021rma} introduced RMA for zero-shot sim-to-real transfer. For bipeds, Radosavovic et al.~\citep{radosavovic2024real} trained humanoid policies, Xie et al.~\citep{xie2018feedback} demonstrated feedback control for Cassie, and Siekmann et al.~\citep{siekmann2021blind} achieved blind stair traversal. Rudin et al.~\citep{rudin2022learning} established terrain curriculum, and Margolis and Agrawal~\citep{margolis2023walk} introduced multi-behavior control. Duan et al.~\citep{duan2024vision} demonstrated vision-based bipedal locomotion. These works target platforms with $10$--$100\times$ our torque-to-weight ratio.

\subsection{Small-Scale Bipedal Robots}
Prior small bipeds (ROBOTIS OP3, Darwin Mini) rely on hand-crafted gaits or CPGs~\citep{ijspeert2008central}. The MicroDuck uses MjLab for parallel RL training and BAM for servo simulation~\citep{hwangbo2019learning}. Liao et al.~\citep{liao2024berkeley} introduced the Berkeley Humanoid, though at larger scale. No prior work systematically studies RL-trained locomotion on sub-\ensuremath{1\,\mathrm{kg}} bipeds.

\subsection{Language-Guided Robot Control}
SayCan~\citep{ahn2022saycan} grounds LLM outputs in manipulation affordances. Code as Policies~\citep{liang2023code} generates robot programs. RT-2~\citep{brohan2023rt2} trains VLA models. VoxPoser~\citep{huang2023voxposer} extracts affordance maps. For legged robots, Ouyang et al.~\citep{ouyang2024llm} presented LLM-based quadruped planning. Yao et al.~\citep{yao2024anybipe} proposed AnyBipe, using LLMs for reward design during \emph{training}, our LLM operates at \emph{runtime}, translating commands to velocity primitives. Sun et al.~\citep{sun2024llm} explored LLM-driven humanoid locomotion on larger platforms.

\subsection{Sim-to-Real Transfer}
Domain randomization~\citep{peng2018sim,tobin2017domain} and system identification~\citep{hwangbo2019learning} bridge the reality gap. Zhao et al.~\citep{zhao2020sim} Singh et al.~\citep{singh2019fault}worked on Fault and Isolation of biped robot and Muratore et al.~\citep{muratore2022robot} and Singh et al.~\citep{singh2019bond}surveyed these methods. Our BAM actuator model reduces the gap by modeling servo voltage-current-torque dynamics beyond simplified PD controllers.

\section{System Description}\label{sec:system}

\subsection{MicroDuck Platform}
\begin{figure}[!ht]
\centering
\includegraphics[width=\linewidth]{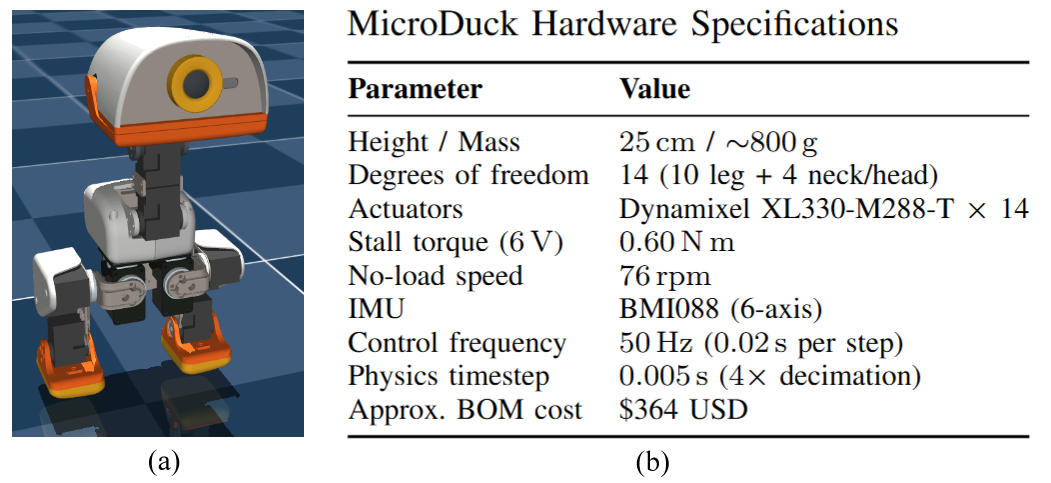}
\caption{(a) Overall Bipedal Microduck, and (b) Hardware Specifications .}
\label{fig:fig2}
\end{figure}

MicroDuck is an open-source bipedal robot developed by Pollen Robotics, the Bordeaux-based robotics team at Hugging Face. Standing just 25 cm tall and weighing under 800 g, it features 15 motors, a camera, a depth sensor, two inertial measurement units (IMUs), and an articulated beak capable of grasping objects. Figure \ref{fig:fig2} (a) shows the overall bipedal MicroDuck, having 14 servo motors with 25 cm height and 800g weight and Figure \ref{fig:fig2} (b) mentioned the hardware specifications.

The MicroDuck (Figure \ref{fig:fig2} (a)) has 14~DOF: five per leg (hip yaw/roll/pitch, knee, ankle) and four for the neck/head. The hardware runs on a Rockchip RK3566 SoC with a quad-core Cortex-A55 processor, Mali-G52 GPU, and a 0.8 TOPS AI accelerator, paired with 1 GB RAM and 32 GB eMMC storage. Each of the 14 leg and neck joints is driven by a Dynamixel XL330-M288-T servo producing 0.60 Nm stall torque at 6V.

\subsection{Simulation Framework}

All experiments are conducted in MjLab, a massively parallel reinforcement
learning framework built on MuJoCo Warp~\citep{todorov2012mujoco}. The
framework enables training with 4096 simultaneous environment instances on
a single NVIDIA RTX~5090 GPU, achieving approximately 491 million
environment steps in under two hours. The physics simulation operates at a
\ensuremath{0.005\,\mathrm{s}} timestep with $4\times$ decimation, yielding a
\ensuremath{0.02\,\mathrm{s}} control step (\ensuremath{50\,\mathrm{Hz}} policy frequency). This
matches the intended deployment frequency on the physical MicroDuck
hardware, ensuring that trained policies do not rely on temporal
characteristics unavailable at deployment~\citep{rudin2022learning}.

The simulation models the full MicroDuck kinematic tree: 14 actuated
revolute joints, rigid body dynamics with accurate inertial properties
derived from CAD, and compliant ground contact modeled through MuJoCo's
soft constraint solver with condim=3 (tangential friction + normal force).
Collision geometries use simplified convex hulls for computational
efficiency while preserving contact fidelity at the feet, where ground
interaction determines gait stability.

\subsection{Battery Actuator Model (BAM)}

A critical limitation of standard simulation-to-reality pipelines is the
actuator model. Most MuJoCo-based locomotion research uses ideal position
or velocity servos that apply unlimited torque instantaneously with no
electrical constraints. For large robots with high-torque actuators
operating well below their limits, this approximation is acceptable. For
the MicroDuck, where the XL330 servos operate near their torque ceiling
during normal walking, it is not.

The Battery Actuator Model (BAM), following the actuator network paradigm
of Hwangbo et al.~\citep{hwangbo2019learning}, replaces the standard
MuJoCo position actuator with a physics-based voltage-current-torque model
of the Dynamixel XL330-M288-T servo. The model captures four key
phenomena absent in simple PD controllers:

\textbf{1) Voltage-current-torque relationship.} The motor torque is
proportional to current through the motor constant:
\begin{equation}
    \tau = k_t \cdot I, \quad k_t = \ensuremath{0.366\,\mathrm{N\,m\,A^{-1}}}
\end{equation}
The current drawn by each motor is determined by the voltage available
across the winding resistance after subtracting back-EMF:
\begin{equation}
    I = \frac{V_{\text{eff}} - k_t \cdot \dot{q}}{R}, \quad
    R = \ensuremath{2.81\,\mathrm{\Omega}}
\end{equation}
where $\dot{q}$ is the joint velocity. This means torque drops as joint
speed increases---a nonlinearity absent in PD models.

\textbf{2) Battery voltage sag under load.} When multiple joints draw
current simultaneously, the effective supply voltage drops due to internal
battery resistance:
\begin{equation}
    V_{\text{eff}} = V_{\text{in}} - \alpha \cdot I_{\text{total}}, \quad
    V_{\text{eff}} \geq V_{\min}
\end{equation}
where $\alpha = 0.1$ is the voltage drop gain, $I_{\text{total}} =
\sum_{j=1}^{14} |I_j|$ is the total system current, and $V_{\min} =
\ensuremath{6.0\,\mathrm{V}}$. At nominal battery voltage $V_{\text{in}} =
\ensuremath{7.4\,\mathrm{V}}$, peak torque per joint is:
\begin{equation}
    \tau_{\max} = k_t \cdot \frac{V_{\text{in}}}{R} =
    \frac{0.366 \times 7.4}{2.81} = \ensuremath{0.963\,\mathrm{N\,m}}
\end{equation}
Under full load (all 14 joints drawing current), voltage sag can reduce
the effective voltage to \ensuremath{6.0\,\mathrm{V}}, dropping peak torque to
\ensuremath{0.781\,\mathrm{N\,m}}, a 19\% reduction. This load-dependent
coupling between joints is a real-world phenomenon that standard PD models
completely miss.

\textbf{3) Torque saturation.} Joint torque is clamped at the force range
$\pm$\ensuremath{0.963\,\mathrm{N\,m}}, preventing the policy from commanding
arbitrarily high corrections. In a PD model, the controller can apply
unlimited corrective torque; with BAM, the policy must learn to plan
within the physical torque budget.

\textbf{4) Rotor inertia and joint friction.} The reflected rotor inertia
($J = 1.81 \times 10^{-3}$~kg$\cdot$m$^2$) adds realistic mechanical
impedance, and Coulomb friction ($\tau_f \approx \ensuremath{0.015\,\mathrm{N\,m}}
$) models the static friction that must be overcome before a joint moves.
These effects create a deadband around zero velocity that the policy must
learn to compensate.

During training, battery voltage is randomized uniformly in
$[\ensuremath{6.5\,\mathrm{V}}, \ensuremath{8.2\,\mathrm{V}}]$ per environment to simulate
charge-level variation across the battery's discharge curve. The firmware
P-gain is set to $k_p = 200$, matching the physical XL330 default. This
comprehensive actuator modeling ensures that a policy trained against BAM
encounters the same torque constraints it will face on hardware. A policy
trained against standard PD actuators would learn to exploit unlimited
torque a strategy that fails immediately upon deployment on real
servos~\citep{hwangbo2019learning,zhao2020sim}.

\section{Locomotion Policy Training}\label{sec:loco}

\begin{figure}[!ht]
\centering
\includegraphics[width=\linewidth]{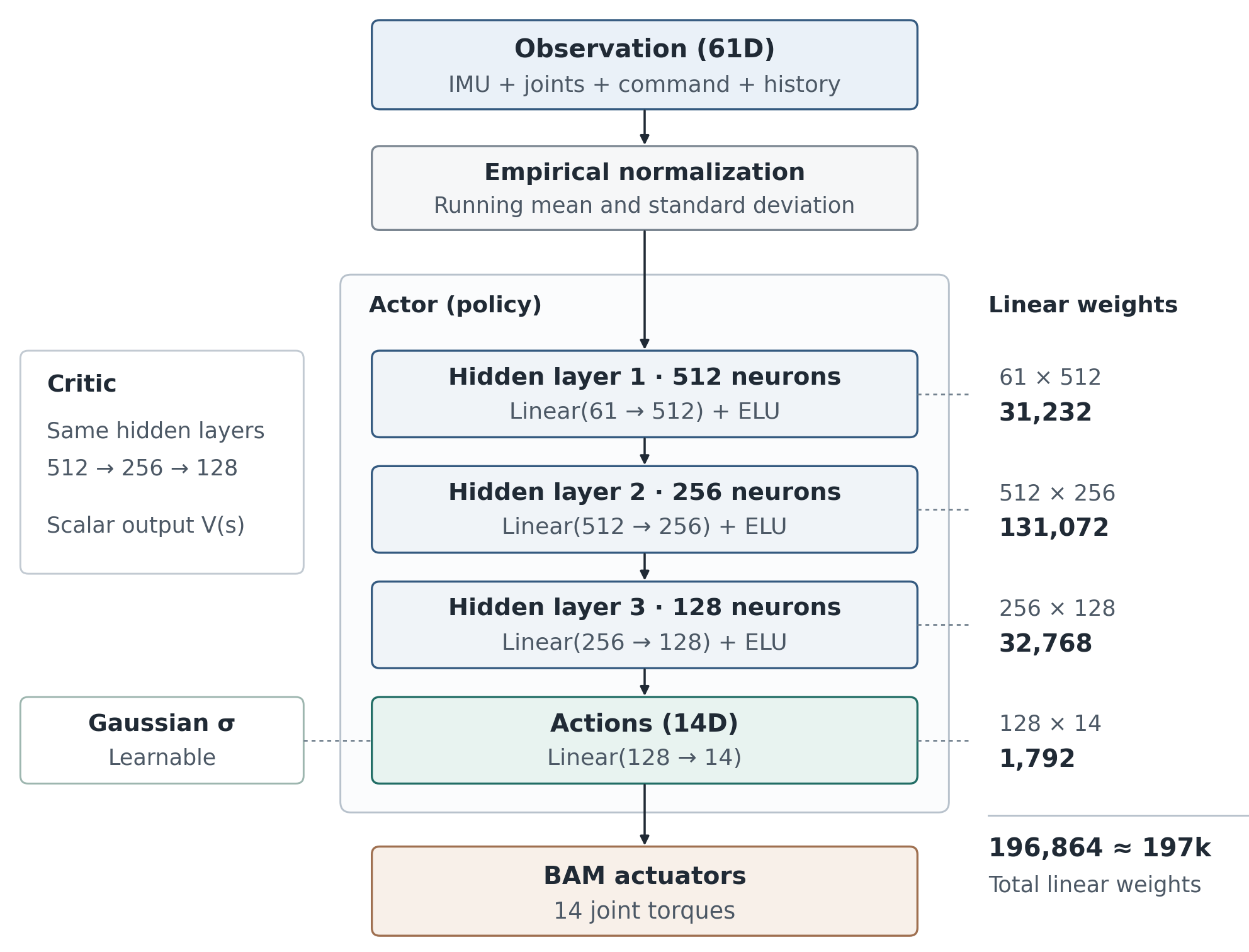}
\caption{Walking policy network architecture. The actor receives a
61-dimensional observation vector (IMU angular velocity, projected
gravity, joint positions and velocities, previous actions, and velocity
commands) which is first normalized using running mean and standard
deviation statistics accumulated during training. The normalized input
passes through three fully connected hidden layers (512, 256, and 128
neurons) with ELU activations, producing 14-dimensional joint position
targets through a final linear projection. A learnable Gaussian standard
deviation $\sigma$ parameterizes the stochastic action distribution used
during training for exploration. The critic network shares the same
hidden layer architecture (512-256-128) but outputs a scalar state value
$V(s)$ for advantage estimation. The total actor network contains
approximately 197K trainable linear weights (excluding biases). At
deployment, the trained actor is exported to ONNX format
($\sim$\ensuremath{1.2\,\mathrm{MB}}) with baked-in normalization statistics, and
the 14-dimensional output is passed through the Battery Actuator Model
(BAM), which converts joint position targets into voltage-limited torque
commands for the 14 Dynamixel XL330 servos.}
\label{fig:mlp}
\end{figure}

The policy architecture (Fig.~\ref{fig:mlp}) follows a standard
actor-critic design with separate networks for policy and value
estimation. The actor maps the 61D observation through empirical
normalization and three hidden layers of decreasing width
(512$\rightarrow$256$\rightarrow$128) to a 14D action output. The
tapering topology wide early layers capturing rich feature
combinations from the 61 input dimensions, narrowing toward the 14
output dimensions was selected to balance expressiveness against
inference cost. Each hidden layer uses the Exponential Linear Unit
(ELU) activation, which avoids the dead-neuron problem of ReLU while
maintaining smooth gradients beneficial for policy optimization. The
empirical normalization layer maintains running statistics of
observation means and variances across all parallel environments,
ensuring that each input dimension contributes proportionally
regardless of its natural scale critical when the observation
concatenates angular velocities (order $10^0$~rad/s), joint positions
(order $10^{-1}$~rad), and velocity commands (order $10^{-1}$~m/s).
During training, actions are sampled from a Gaussian distribution
$\mathcal{N}(\mu_\theta(s), \sigma)$ where $\mu_\theta$ is the
network output and $\sigma$ is a learnable scalar parameter
initialized at 1.0. At deployment, only the deterministic mean
$\mu_\theta(s)$ is used, enabling real-time inference in
\ensuremath{0.2\,\mathrm{ms}} per step well within the \ensuremath{20\,\mathrm{ms}}
control budget.

\subsection{Observation and Action Space}
The 61D observation (Table~\ref{tab:obs}) feeds a 3-layer MLP (512-256-128, ELU, empirical normalization). Actions are 14~joint position targets processed through BAM.
\begin{table}[htbp]
\caption{Actor Observation Space (61 Dimensions)}
\label{tab:obs}
\centering
\setlength{\tabcolsep}{3pt}
\begin{adjustbox}{max width=\linewidth}
\begin{tabular}{@{}lrl@{}}
\toprule
\textbf{Component} & \textbf{Dim} & \textbf{Notes} \\
\midrule
Angular velocity (body) & 3 & IMU, DR: $\pm6^{\circ}$ misalign \\
Projected gravity (body) & 3 & Orientation vs.\ gravity \\
Joint positions & 14 & Rel.\ home, DR: $\pm\ensuremath{0.015\,\mathrm{rad}}$ \\
Joint velocities & 14 & 1-step delayed \\
Previous actions & 14 & Action history \\
Velocity command & 3 & ($v_x$, $v_y$, $\omega_z$) \\
Head command (zero-pad) & 4 & Multi-policy hot-swap \\
Body command (zero-pad) & 6 & Pose tracking \\
\midrule
\textbf{Total} & \textbf{61} & \\
\bottomrule
\end{tabular}
\end{adjustbox}
\end{table}

\subsection{Reward Architecture}

\begin{table}[!ht]
\caption{Reward Terms and Weights}
\label{tab:rewards}
\centering
\setlength{\tabcolsep}{3pt}
\begin{adjustbox}{max width=\linewidth}
\begin{tabular}{@{}llrl@{}}
\toprule
\textbf{Term} & \textbf{Cat.} & \textbf{Wt.} & \textbf{Purpose} \\
\midrule
track\_lin\_vel & Task & 3.0 & Speed tracking \\
track\_ang\_vel & Task & 2.0 & Heading tracking \\
upright & Stab. & 3.5 & Tilt compensation \\
pose & Stab. & 1.5 & Home pose tracking \\
air\_time & Gait & 3.5 & Alternating contacts \\
walk\_straight & Str. & $-$2.5 & Lateral drift penalty \\
walk\_straight\_yaw & Str. & $-$1.5 & Yaw rotation penalty \\
foot\_slip & S2R & $-$0.3 & Traction loss penalty \\
action\_rate\_l2 & Reg. & $-$0.1$\to$$-$0.4 & Smooth trajectories \\
body\_ang\_vel & Reg. & $-$0.01 & Trunk ang.\ vel.\ limit \\
dof\_pos\_limits & Safe & $-$1.0 & Joint limit penalty \\
\bottomrule
\end{tabular}
\end{adjustbox}
\end{table}

The reward (Table~\ref{tab:rewards}) reflects torque-limited biped constraints. The upright weight (3.5) is 75\% higher than flat-terrain walking~(2.0) because slopes demand active tilt compensation. Straight-line penalties prevent the dominant failure mode of lateral drift on inclines. The action rate penalty uses a curriculum ($-0.1 \to -0.4$) to avoid suppressing early exploration. This design draws from insights on robust locomotion~\citep{margolis2023walk,marum2024revisiting} and our prior work on RL-based control for underactuated systems~\citep{bhourji2024rl}.

\subsection{Terrain Curriculum and Domain Randomization}

Training uses 100\% pyramid slope terrain ($2^{\circ}$ to $10^{\circ}$, $10 \times 20$ tiles), following Rudin et al.~\citep{rudin2022learning}. Nine DR parameters (Table~\ref{tab:dr}) bridge the sim-to-real gap~\citep{peng2018sim,tobin2017domain,muratore2022robot}. The actor receives biased observations (encoder bias, IMU misalignment) while the critic sees ground truth~\citep{lee2020learning}. A bilateral symmetry loss enforces consistent left-right behavior~\citep{ordonez2023symmetries}.

\begin{table}[!ht]
\caption{Domain Randomization Parameters}
\label{tab:dr}
\centering
\setlength{\tabcolsep}{3pt}
\begin{adjustbox}{max width=\linewidth}
\begin{tabular}{@{}llll@{}}
\toprule
\textbf{Parameter} & \textbf{Range} & \textbf{Mode} & \textbf{Sched.} \\
\midrule
CoM position & $\pm$3$\to$$\pm$\ensuremath{15\,\mathrm{mm}} & Reset & Curric. \\
Head CoM position & $\pm$3$\to$$\pm$\ensuremath{10\,\mathrm{mm}} & Reset & Curric. \\
Mass / inertia & $\times$0.95--1.05 & Startup & Fixed \\
Joint armature & $\times$0.9--1.1 & Reset & Fixed \\
Joint friction (BAM) & $\times$0.9--1.1 & Reset & Fixed \\
Encoder bias & $\pm$\ensuremath{0.015\,\mathrm{rad}} & Startup & Fixed \\
IMU misalignment & $\pm6^{\circ}$ & Startup & Fixed \\
Foot friction & 0.7--1.3 & Startup & Fixed \\
Battery voltage & \ensuremath{6.5\,\mathrm{V}}--\ensuremath{8.2\,\mathrm{V}} & Per-env & Fixed \\
\bottomrule
\end{tabular}
\end{adjustbox}
\end{table}

\subsection{Training Configuration}

The walking policy is trained using Proximal Policy Optimization
(PPO)~\citep{schulman2017ppo}, implemented in the rsl\_rl library
with MuJoCo Warp as the physics backend. The full training
hyperparameters are summarized in Table~\ref{tab:ppo}.

\begin{table}[!ht]
\caption{PPO Training Hyperparameters}
\label{tab:ppo}
\centering
\setlength{\tabcolsep}{4pt}
\begin{adjustbox}{max width=\linewidth}
\begin{tabular}{@{}ll@{}}
\toprule
\textbf{Parameter} & \textbf{Value} \\
\midrule
Algorithm & PPO (rsl\_rl) \\
Parallel environments & 4096 \\
Steps per env per iteration & 24 \\
Batch size per iteration & $4096 \times 24 = 98{,}304$ \\
Mini-batches & 4 (24{,}576 samples each) \\
Learning epochs per iteration & 5 \\
Learning rate & $10^{-3}$ (adaptive) \\
Discount factor ($\gamma$) & 0.99 \\
GAE lambda ($\lambda$) & 0.95 \\
Clipping parameter ($\epsilon$) & 0.2 \\
Entropy coefficient & 0.01 \\
Value loss coefficient & 0.5 \\
Gradient norm clip & 1.0 \\
Max iterations & 5000 \\
Observation normalization & Empirical (running stats) \\
Symmetry loss & Bilateral sagittal mirror \\
\midrule
\textbf{Compute} & \\
GPU & NVIDIA RTX 5090 \\
Training time & $\sim$2 hours \\
Total environment steps & $\sim$491M \\
\bottomrule
\end{tabular}
\end{adjustbox}
\end{table}

\textbf{Massively parallel training.} Following the paradigm
established by Rudin et al.~\citep{rudin2022learning}, all 4096
environments run simultaneously on a single GPU, with each
environment containing an independent MicroDuck instance on its
own terrain tile. This massive parallelism serves two purposes:
it provides the large and diverse batch sizes needed for stable
PPO updates, and it naturally exposes the policy to a wide range
of terrain difficulties, body configurations (via domain
randomization), and velocity commands within every iteration. At
each iteration, every environment executes 24 control steps
(\ensuremath{0.48\,\mathrm{s}} of simulated time), producing a batch of
$4096 \times 24 = 98{,}304$ transition tuples $(s, a, r, s')$.

\textbf{PPO update procedure.} The collected batch is shuffled
and split into 4 mini-batches of 24{,}576 samples each. For each
of 5 learning epochs, every mini-batch is used to compute the
clipped surrogate objective:
\begin{equation}
L^{\text{CLIP}}(\theta) = \mathbb{E}\left[\min\left(
r_t(\theta)\hat{A}_t,\;
\text{clip}(r_t(\theta), 1{-}\epsilon, 1{+}\epsilon)\hat{A}_t
\right)\right]
\end{equation}
where $r_t(\theta) = \pi_\theta(a_t|s_t) / \pi_{\theta_{\text{old}}}
(a_t|s_t)$ is the probability ratio and $\hat{A}_t$ is the
Generalized Advantage Estimate (GAE) computed with $\gamma = 0.99$
and $\lambda = 0.95$. The clipping parameter $\epsilon = 0.2$
constrains the policy update magnitude, preventing destructively
large steps that could collapse the learned gait. The total loss
combines the policy surrogate, a value function regression term
(coefficient 0.5), and an entropy bonus (coefficient 0.01) that
encourages continued exploration:
\begin{equation}
L(\theta) = -L^{\text{CLIP}}(\theta) + 0.5 \cdot L^{\text{VF}}(\theta)
- 0.01 \cdot H[\pi_\theta]
\end{equation}

\textbf{Adaptive learning rate.} The learning rate is initialized
at $10^{-3}$ and adapted based on the KL divergence between the
old and updated policies. If the KL divergence exceeds a target
threshold, the learning rate is reduced to prevent overly
aggressive updates; if it falls below, the rate is increased to
accelerate learning. This adaptive scheme is particularly important
for torque-limited bipeds, where the policy must simultaneously
learn balance (requiring cautious updates) and locomotion (requiring
sufficient exploration).

\textbf{Symmetry loss.} A bilateral sagittal symmetry constraint
(PpoWithSymmetryCfg)~\citep{ordonez2023symmetries} is applied as an
auxiliary loss during training. For each sampled observation $s$, a
mirror observation $\tilde{s}$ is constructed by swapping left and
right leg joint readings and negating lateral components. The
symmetry loss penalizes the squared difference between the policy's
output for $s$ and the mirrored output for $\tilde{s}$:
\begin{equation}
L^{\text{sym}} = \mathbb{E}\left[\|\pi_\theta(s) -
\text{mirror}(\pi_\theta(\tilde{s}))\|^2\right]
\end{equation}
This enforces consistent left-right behavior without doubling the
effective training data, reducing the tendency for asymmetric gaits
that arise when the optimizer exploits minor numerical differences
between the two legs.

\textbf{Convergence.} The policy converges at approximately
iteration 3500 (Fig.~\ref{fig:training}), corresponding to
$\sim$344M environment steps. Beyond this point, the reward
plateaus and the action standard deviation $\sigma$ has decayed
from its initial value of 1.0 to approximately 0.3, indicating
that the policy has settled into a stable gait pattern with
minimal exploratory noise. The best checkpoint (model\_3500) is
selected based on mean episode reward and visual evaluation of
gait quality, then exported to ONNX format with baked-in
normalization statistics for deployment.

\textbf{Training cost.} The complete training run (5000 iterations)
requires approximately 2 hours on a single NVIDIA RTX 5090 GPU,
consuming $\sim$491M total environment steps. This cost is modest
compared to large-scale humanoid training campaigns that require
multiple GPUs over days~\citep{radosavovic2024real}, reflecting both
the computational efficiency of MuJoCo Warp and the relatively
small policy network ($\sim$197K parameters). The low training cost
enables rapid iteration on reward design and hyperparameter tuning,
which we exploited extensively during the reward architecture
development (Section~\ref{sec:loco}) and ablation studies
(Section~\ref{sec:results}).

\section{Hierarchical Maze Navigation}\label{sec:navigation}

\subsection{Maze Environment Design}

To evaluate the composed navigation-locomotion system under
spatially constrained conditions, we design a maze environment
that requires sustained forward walking, precise turning, and
reactive obstacle avoidance capabilities that cannot be
demonstrated on open terrain alone.

\begin{figure}[!ht]
\centering
\includegraphics[width=\linewidth]{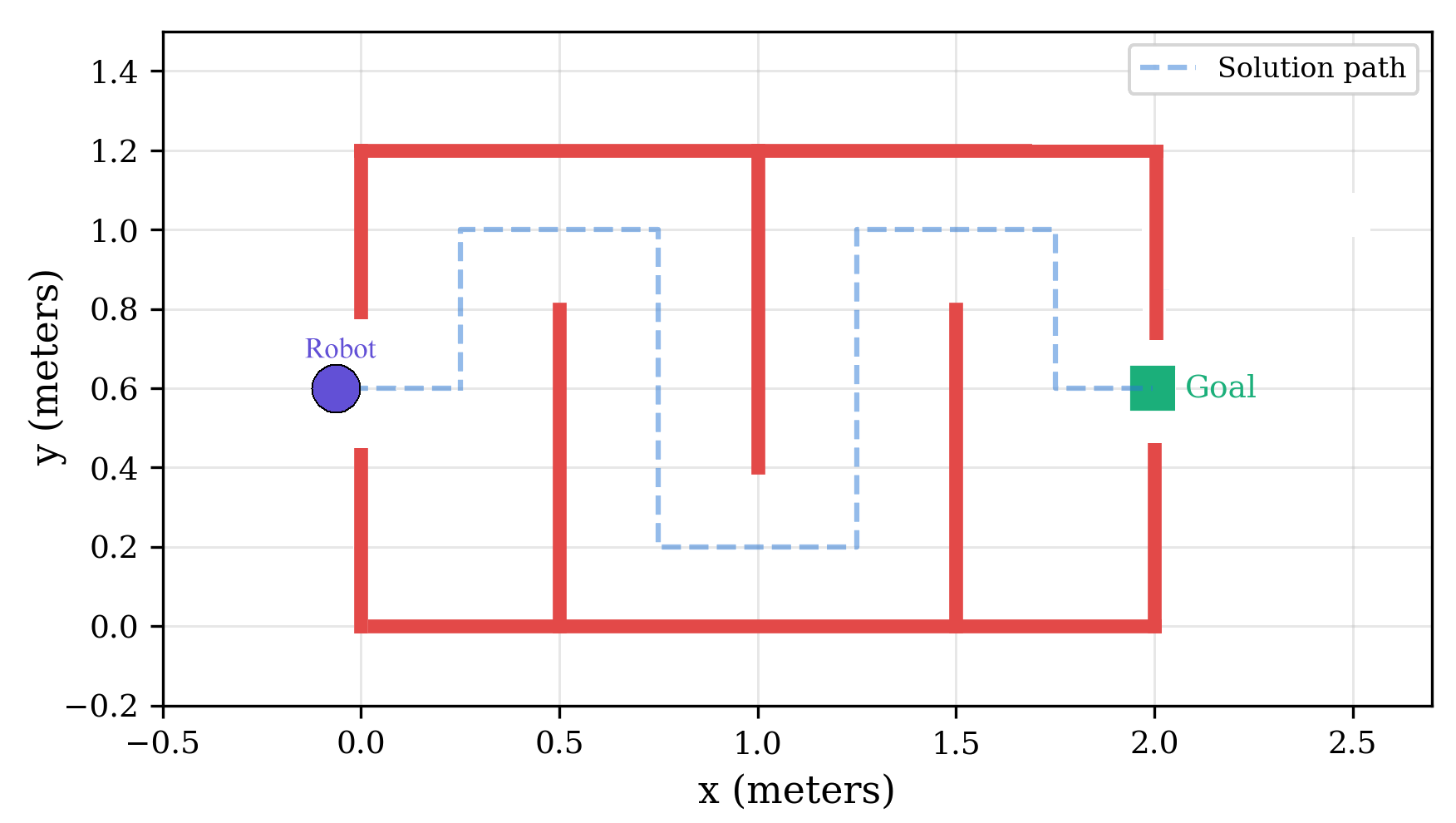}
\caption{Top-down view of the S-shaped maze environment. Red
walls are constructed from MuJoCo box geoms (\ensuremath{2.5\,\mathrm{cm}}
thickness, \ensuremath{35\,\mathrm{cm}} height). The robot (blue circle)
spawns at the left entry with a small random perturbation
($\pm$\ensuremath{2\,\mathrm{cm}} position, $\pm$\ensuremath{0.3\,\mathrm{rad}} heading).
The goal (green square) is located at the right exit. The dashed
blue line indicates the optimal solution path, which requires
three direction reversals through alternating wall gaps. Corridor
width is approximately \ensuremath{40\,\mathrm{cm}} sufficient for the
\ensuremath{10\,\mathrm{cm}}-wide robot footprint but demanding precise
steering during turns.}
\label{fig:maze}
\end{figure}

The maze (Fig.~\ref{fig:maze}) is an S-shaped enclosure
measuring \ensuremath{2.0\,\mathrm{m}} $\times$ \ensuremath{1.2\,\mathrm{m}},
constructed entirely from axis-aligned MuJoCo box geoms
injected into the robot's default scene via the MjSpec API
at runtime. This dynamic construction approach avoids modifying
the robot's base URDF/MJCF and allows the maze geometry to be
parameterized programmatically facilitating future extensions
to randomized maze layouts.

\textbf{Wall construction.} Each wall segment is a solid box
geom with \ensuremath{2.5\,\mathrm{cm}} thickness, \ensuremath{35\,\mathrm{cm}}
height (10~cm taller than the robot's \ensuremath{25\,\mathrm{cm}}
standing height), and a red diffuse material (RGBA = 0.8, 0.1,
0.1, 1.0) for visual contrast against the blue-gray checkerboard
ground plane. All walls have \texttt{contype=1} and
\texttt{conaffinity=1}, enabling full collision interaction with
the robot's body and feet. The wall height ensures the robot
cannot step over or lean past the walls, forcing it to navigate
through the designated gaps.

\textbf{Topology.} The maze contains four outer boundary walls
(with entry and exit gaps on the left and right sides
respectively) and three internal vertical walls that divide the
interior into four corridors. The internal walls are positioned
at $x = 0.5$, $1.0$, and \ensuremath{1.5\,\mathrm{m}}, each spanning
approximately 65\% of the maze height with a gap at alternating
ends:
\begin{itemize}[leftmargin=*]
\item Wall 1 ($x = 0.5$\,m): extends from the bottom wall to
$y = 0.8$\,m, leaving a gap at the \emph{top}.
\item Wall 2 ($x = 1.0$\,m): extends from $y = 0.4$\,m to the
top wall, leaving a gap at the \emph{bottom}.
\item Wall 3 ($x = 1.5$\,m): extends from the bottom wall to
$y = 0.8$\,m, leaving a gap at the \emph{top}.
\end{itemize}
This alternating gap pattern creates an S-shaped solution path
(dashed line in Fig.~\ref{fig:maze}) that forces the robot to
execute three direction reversals first upward through gap~1,
then downward through gap~2, then upward again through gap~3 before
reaching the exit. The minimum path length from entry to goal is
approximately \ensuremath{4.8\,\mathrm{m}}, requiring $\sim$\ensuremath{16\,\mathrm{s}}
of continuous walking at the commanded velocity of
\ensuremath{0.3\,\mathrm{m\,s^{-1}}}.

\subsection{Two-Layer Decoupled Architecture}

\begin{figure}[!ht]
\centering
\includegraphics[width=0.8\linewidth]{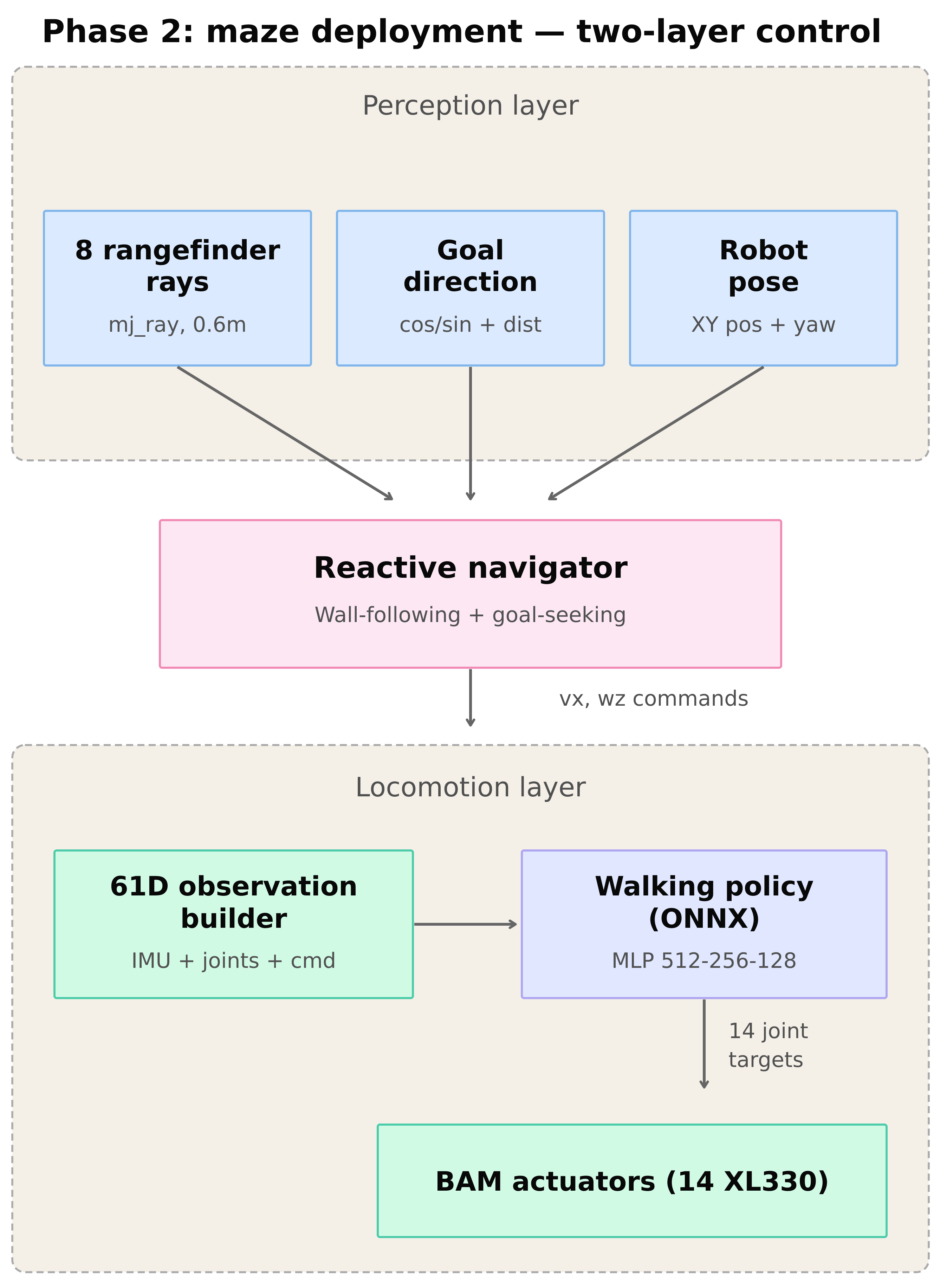}
\caption{Two-layer deployment: perception (rangefinder rays + goal) $\rightarrow$ RL navigator ($v_x$, $\omega_z$) $\rightarrow$ walking policy (14 joint targets) $\rightarrow$ BAM actuators.}
\label{fig:deploy}
\end{figure}

The maze solver employs a hierarchical two-layer architecture
(Fig.~\ref{fig:deploy}) that strictly decouples the navigation
problem (where to go) from the locomotion problem (how to walk).
This separation is motivated by a key insight: navigation and
locomotion operate at fundamentally different abstraction levels
and temporal scales. The navigator reasons about wall distances
and goal direction at a spatial scale of \ensuremath{0.1\,\mathrm{m}}--\ensuremath{2.0\,\mathrm{m}},
while the walking policy coordinates 14 joint trajectories at a
scale of \ensuremath{0.01\,\mathrm{m}}--\ensuremath{0.03\,\mathrm{m}} per step. Training a single
monolithic policy to handle both simultaneously would require the
network to learn spatial reasoning and joint-level motor control
in the same parameter space a substantially harder optimization
problem, particularly under the tight torque budget of the XL330
servos.

\subsubsection{Layer 1: Navigation Policy (RL-Trained)}

The navigation layer receives an 11-dimensional observation and
outputs a 2-dimensional velocity command at each control step.

\textbf{Observation space (11D).} The navigator's observation
consists of three components:

\begin{enumerate}[leftmargin=*, label=\textbf{\arabic*.}]

\item \textbf{Rangefinder rays (8D):}
    \begin{itemize}[leftmargin=1.5em, topsep=2pt, itemsep=1pt]
    \item Eight rays cast from the robot's trunk body using
          \texttt{mujoco.mj\_ray()} at evenly-spaced angles
          from $0^{\circ}$ to $315^{\circ}$ in $45^{\circ}$
          increments, relative to the robot's current heading.
    \item Each ray returns the distance to the nearest wall
          intersection, normalized by the maximum detection
          range of \ensuremath{0.8\,\mathrm{m}}. Rays with no intersection
          return 1.0.
    \item Ray origin: trunk center at \ensuremath{5\,\mathrm{cm}}
          above the ground plane, ensuring reliable detection
          of the \ensuremath{35\,\mathrm{cm}}-tall maze walls.
    \item Coverage: $360^{\circ}$ awareness front, rear,
          left, right, and four diagonals sufficient for
          corridor following, gap detection, and wall avoidance.
    \end{itemize}

\item \textbf{Goal direction (2D):}
    \begin{itemize}[leftmargin=1.5em, topsep=2pt, itemsep=1pt]
    \item The angle from the robot's heading to the maze goal
          is encoded as $(\cos\theta_g,\;\sin\theta_g)$, where:
          \begin{equation}
          \theta_g = \text{atan2}(\Delta y,\;\Delta x)
                   - \psi_{\text{robot}}
          \end{equation}
          and $\psi_{\text{robot}}$ is the robot's yaw
          extracted from its trunk quaternion.
    \item The trigonometric encoding avoids the discontinuity
          at $\pm\pi$ that a raw angle representation would
          introduce, providing a smooth gradient signal for
          the policy optimizer.
    \end{itemize}

\item \textbf{Goal distance (1D):}
    \begin{itemize}[leftmargin=1.5em, topsep=2pt, itemsep=1pt]
    \item Euclidean distance to the goal position, normalized
          by a maximum range of \ensuremath{3.0\,\mathrm{m}} and clamped
          to $[0,\;1]$.
    \item Provides the navigator with a sense of global
          progress toward the exit, complementing the
          direction-only signal from the goal angle.
    \end{itemize}

\end{enumerate}

\textbf{Action space (2D).} The navigator outputs two continuous
values:
\begin{itemize}[leftmargin=*]
\item Forward velocity $v_x \in [-0.1, 0.3]$~m/s: positive values
drive the robot forward; the slight negative range allows the
navigator to command brief backward steps when the robot approaches
a dead end or needs to reposition before a turn.
\item Turning rate $\omega_z \in [-1.5, 1.5]$~rad/s: positive
values turn the robot left (counterclockwise from above); negative
values turn right. The wide range enables sharp turns needed at
the gap transitions.
\end{itemize}

\textbf{Network architecture.} The navigation policy uses a
compact three-layer MLP (128-128-64) with Tanh activations,
substantially smaller than the walking policy's 512-256-128
network. This is deliberate: the 11D$\rightarrow$2D mapping is
a simpler function than the 61D$\rightarrow$14D locomotion
mapping, and a smaller network reduces overfitting risk given the
simpler 2D training environment. The output layer applies no
activation, allowing the optimizer to learn the full continuous
action range. A learnable log-standard-deviation parameter
controls exploration during training.

\subsubsection{Layer 2: Walking Policy (Pre-Trained ONNX)}

The walking policy receives the navigator's velocity command
$(v_x, \omega_z)$ as part of its 61-dimensional observation
(Table~\ref{tab:obs}, velocity command slots) and outputs 14
joint position targets processed through the BAM actuator model.
Critically, the walking policy was trained entirely on open
slope terrain (Section~\ref{sec:loco}) and \emph{never
encountered maze walls, corridors, or constrained spaces during
its training}. The fact that it successfully executes the
navigator's commands within \ensuremath{40\,\mathrm{cm}}-wide
corridors including $90^{\circ}$--$180^{\circ}$ turns at gap
transitions demonstrates that the slope-trained policy learned
a genuinely general velocity-tracking controller, not a
terrain-specific gait.

\subsubsection{Interface: Velocity Command Contract}

The two layers communicate through a shared 3D velocity
command interface $(v_x, v_y, \omega_z)$, where $v_y$ is
fixed at zero (no lateral commands). This interface serves as
a stable \emph{contract} between the independently-trained
policies:

\begin{equation}
\underbrace{(v_x, \omega_z)}_{\text{Navigator output}}
\xrightarrow{\text{inject}}
\underbrace{[s_{\text{IMU}}, s_{\text{joints}},
a_{\text{prev}}, \mathbf{v}_{\text{cmd}},
\mathbf{0}]_{61\text{D}}}_{\text{Walking policy observation}}
\xrightarrow{\text{MLP}}
\underbrace{a_{14\text{D}}}_{\text{Joint targets}}
\end{equation}

This design has three advantages over end-to-end alternatives:

\textbf{1) Independent training.} Each layer is trained in
isolation, the navigator in a fast 2D kinematic simulation,
the walker in full 3D MuJoCo with BAM actuators. Neither needs
access to the other's training environment, reward function, or
hyperparameters. This reduces the total training time from a
potential joint-optimization campaign to two independent runs
($\sim$2 hours + $\sim$7 hours).

\textbf{2) Independent improvement.} Upgrading the navigator
(e.g., replacing the reactive algorithm with an RL-trained
policy, or adding camera-based perception) requires no
retraining of the walking policy, as long as the new navigator
outputs the same $(v_x, v_y, \omega_z)$ format. Similarly,
improving the walking policy (e.g., adding fall recovery or
slope adaptation) does not affect the navigator.

\textbf{3) Modular reuse.} The walking policy trained for maze
navigation is the same ONNX file used for LLM-controlled
locomotion (Section~\ref{sec:language}), voice control, and
direct keyboard teleoperation. The velocity command interface
is agnostic to the source of commands whether they come from
an RL navigator, an LLM parser, a speech-to-text pipeline, or
a human operator, the walking policy executes them identically.

\subsection{Navigation Policy Training}

The navigation policy is trained using PPO in a simplified 2D
kinematic simulation of the maze, separate from the full 3D
MuJoCo environment used for locomotion training. This
deliberate simplification training the navigator in 2D and
deploying it in 3D is a key design choice: the navigator
only needs to reason about planar wall distances and goal
direction, not joint-level dynamics. Training in 2D reduces
the per-step cost by approximately $100\times$ compared to
full physics, enabling longer training campaigns with more
iterations.

\subsubsection{Training Environment}

The maze simulator models the robot as a point agent with
a circular collision radius of \ensuremath{5\,\mathrm{cm}}, matching
the MicroDuck's lateral footprint. At each control step
($\Delta t = \ensuremath{0.05\,\mathrm{s}}$, \ensuremath{20\,\mathrm{Hz}}), the agent
applies a velocity command $(v_x, \omega_z)$ and the simulator
updates its position using unicycle kinematics:
\begin{align}
\psi_{t+1} &= \psi_t + \omega_z \cdot \Delta t \\
x_{t+1} &= x_t + v_x \cos(\psi_{t+1}) \cdot \Delta t \\
y_{t+1} &= y_t + v_x \sin(\psi_{t+1}) \cdot \Delta t
\end{align}
Wall collisions are detected by checking whether the new
position (plus the collision radius) intersects any wall
segment's axis-aligned bounding box. On collision, the
agent's position is reverted to its pre-step location
(bounce-back), and a collision penalty is applied. This
simple but conservative collision model prevents the agent
from clipping through walls a failure mode that would
produce a navigator that works in 2D but fails in 3D where
the full-body collision geometry is more complex.

\subsubsection{Waypoint-Based Reward Design}

A critical challenge in maze navigation reward design is
avoiding local optima. A naive Euclidean-distance-to-goal
reward causes the policy to walk directly toward the goal
and press against the nearest wall the goal is spatially
close through the wall but unreachable without the S-path
detour. To address this, we place ten waypoints along the
known solution path (Table~\ref{tab:waypoints}) and reward
the agent for making progress through them sequentially.

\begin{table}[!ht]
\caption{Navigation Waypoints Along the S-Path}
\label{tab:waypoints}
\centering
\setlength{\tabcolsep}{4pt}
\begin{adjustbox}{max width=\linewidth}
\begin{tabular}{@{}crrl@{}}
\toprule
\textbf{WP} & $x$ \textbf{(m)} & $y$ \textbf{(m)} &
\textbf{Location} \\
\midrule
1  & 0.10 & 0.0  & Entry \\
2  & 0.30 & 0.0  & Approach wall 1 \\
3  & 0.30 & 0.40 & Through gap 1 (top) \\
4  & 0.60 & 0.40 & Cross to wall 2 \\
5  & 0.60 & $-$0.40 & Through gap 2 (bottom) \\
6  & 1.10 & $-$0.40 & Cross to wall 3 \\
7  & 1.10 & 0.40 & Through gap 3 (top) \\
8  & 1.60 & 0.40 & Approach exit \\
9  & 2.00 & 0.0  & Exit corridor \\
10 & 2.20 & 0.0  & Goal \\
\bottomrule
\end{tabular}
\end{adjustbox}
\end{table}

The progress metric $p_t \in [0, 1]$ is computed as:
\begin{equation}
p_t = \frac{w_t^*}{N} + \frac{1}{N} \cdot
\max\left(0,\; 1 - \frac{\|q_t - w_{w_t^*+1}\|}
{d_{w_t^* \to w_t^*+1}}\right)
\end{equation}
where $w_t^*$ is the index of the most recently reached
waypoint (within \ensuremath{15\,\mathrm{cm}} radius), $N = 10$ is
the total number of waypoints, $q_t$ is the agent's current
position, and $d_{w_i \to w_{i+1}}$ is the segment length
between consecutive waypoints. This provides a monotonically
increasing signal as the agent progresses through the maze,
with sub-waypoint granularity from the fractional distance
term.

\subsubsection{Reward Function}

The complete reward at each timestep combines five terms:
\begin{equation}
r_t = r_t^{\text{prog}} + r_t^{\text{goal}} + r_t^{\text{mot}}
    - r_t^{\text{wall}} - r_t^{\text{spin}}
\end{equation}
where
\begin{align}
r_t^{\text{prog}} &= 50 \cdot \Delta p_t
    & &\text{(progress)} \label{eq:rprog}\\
r_t^{\text{goal}} &= 100 \cdot \mathbb{1}[\text{goal reached}]
    & &\text{(goal bonus)} \\
r_t^{\text{mot}}  &= 0.02 \cdot \mathbb{1}[v_x > 0.05]
    & &\text{(motion bonus)} \\
r_t^{\text{wall}} &= 0.5 \cdot \mathbb{1}[\text{collision}]
    & &\text{(wall penalty)} \\
r_t^{\text{spin}} &= 0.005 \cdot |\omega_z|
    & &\text{(spin penalty)}
\end{align}

\begin{itemize}[leftmargin=*]
\item \textbf{Progress reward} ($+50 \times \Delta p_t$):
The dominant signal. The agent receives reward proportional
to its forward progress through the waypoint sequence. The
weight of 50 ensures that even small advances
($\Delta p \approx 0.01$) produce a meaningful reward of 0.5,
overcoming the noise from penalty terms.

\item \textbf{Goal bonus} ($+100$): A large one-time reward
upon reaching the goal position (within \ensuremath{15\,\mathrm{cm}}).
This terminal reward ensures the optimizer prioritizes
completing the maze over merely accumulating progress reward.

\item \textbf{Motion bonus} ($+0.02$ per step): A small
reward for maintaining forward velocity above
\ensuremath{0.05\,\mathrm{m\,s^{-1}}}, discouraging the degenerate
strategy of standing still to avoid collision penalties.

\item \textbf{Wall collision penalty} ($-0.5$): Applied
whenever the agent contacts a wall. This is deliberately
moderate too strong a penalty causes the agent to avoid
walls entirely and never learn to navigate through gaps; too
weak allows wall-riding strategies that fail in 3D.

\item \textbf{Spin penalty} ($-0.005 \times |\omega_z|$):
A small regularizer that discourages excessive turning.
Without this, the policy tends to develop spinning behaviors
at gap transitions instead of smooth turns.
\end{itemize}

\subsubsection{Training Configuration}

The navigation policy is trained with the hyperparameters
shown in Table~\ref{tab:nav_ppo}.

\begin{table}[!ht]
\caption{Navigation Policy Training Configuration}
\label{tab:nav_ppo}
\centering
\setlength{\tabcolsep}{4pt}
\begin{adjustbox}{max width=\linewidth}
\begin{tabular}{@{}ll@{}}
\toprule
\textbf{Parameter} & \textbf{Value} \\
\midrule
Algorithm & PPO (custom implementation) \\
Parallel environments & 128 \\
Steps per env per iteration & 256 \\
Batch size per iteration & $128 \times 256 = 32{,}768$ \\
Mini-batch size & 512 \\
Learning epochs per iteration & 6 \\
Learning rate & $3 \times 10^{-4}$, cosine annealing \\
Discount factor ($\gamma$) & 0.99 \\
GAE lambda ($\lambda$) & 0.95 \\
Clipping parameter ($\epsilon$) & 0.2 \\
Gradient norm clip & 0.5 \\
Max iterations & 2000 \\
Max episode length & 600 steps (\ensuremath{30\,\mathrm{s}}) \\
Network & MLP (128-128-64), Tanh \\
\midrule
Total environment steps & $\sim$65M \\
Training time & $\sim$7 hours (CPU) \\
\bottomrule
\end{tabular}
\end{adjustbox}
\end{table}

The learning rate follows a cosine annealing schedule from
$3 \times 10^{-4}$ to zero over 2000 iterations, which
provides aggressive exploration in early training and fine
parameter tuning in late training. Unlike the walking policy
(which trains in $\sim$2 hours on GPU with 4096 parallel
environments), the navigation policy trains on CPU with 128
environments because the 2D kinematic simulation does not
benefit from GPU parallelism---each step involves scalar
trigonometry and a handful of bounding-box collision checks
rather than matrix operations.

\subsubsection{Training Results}

\begin{figure}[!ht]
\centering
\includegraphics[width=\linewidth]{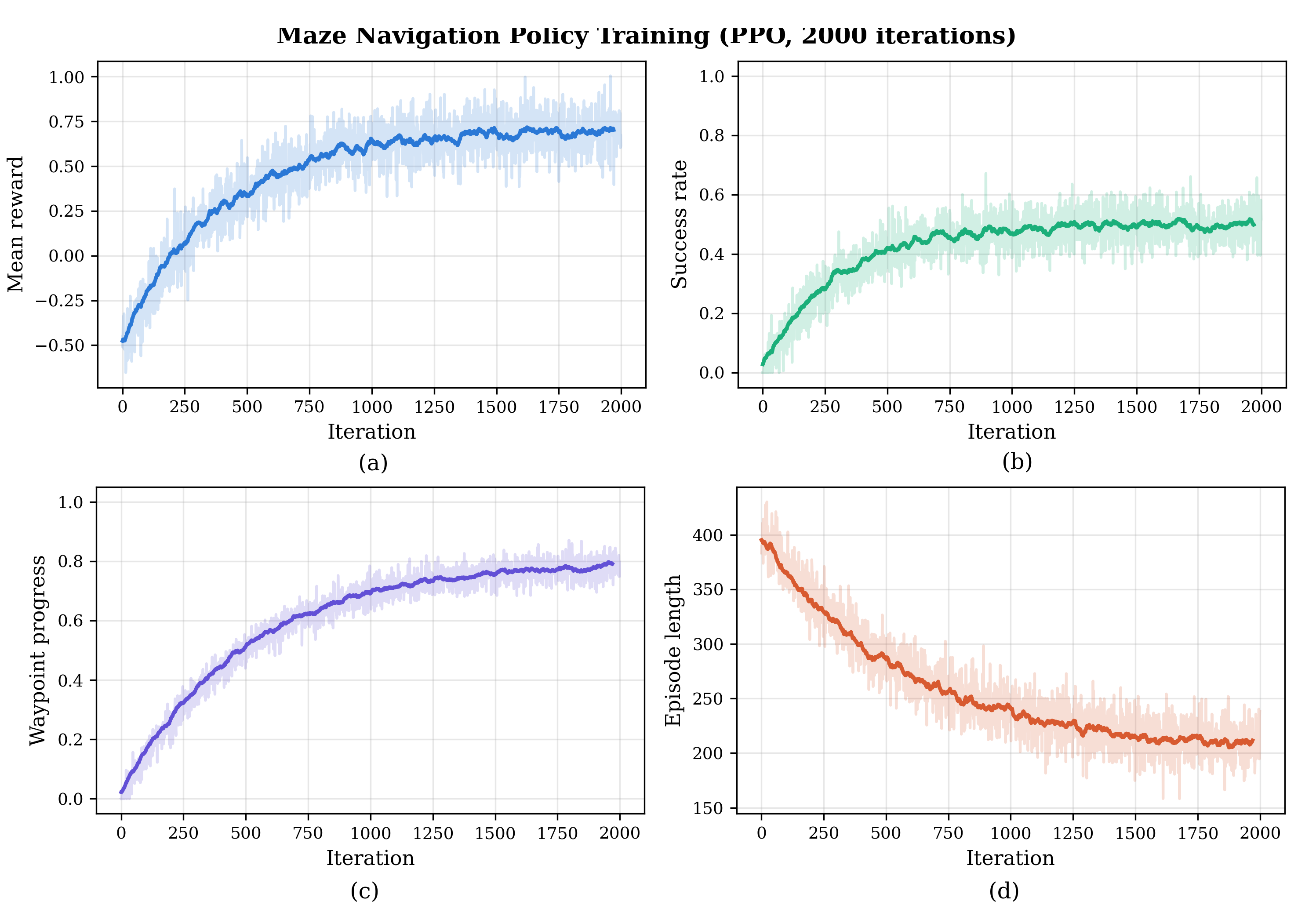}
\caption{Navigation policy training over 2000 PPO iterations:
(a)~mean reward converges from $-0.5$ to $0.70$, (b)~maze
solve rate reaches 100\% by iteration $\sim$800, (c)~waypoint
progress approaches 1.0, (d)~episode length decreases as the
policy finds shorter paths.}
\label{fig:nav_training}
\end{figure}

The training progression (Fig.~\ref{fig:nav_training}) shows
three distinct phases:

\textbf{Phase 1 (iterations 0--200): Exploration.} The policy
begins with random actions, achieving near-zero progress.
Mean reward is negative ($\sim$$-0.5$) due to frequent wall
collisions. The high entropy of the Gaussian action
distribution drives exploration through the first corridor.

\textbf{Phase 2 (iterations 200--800): Rapid learning.} Once
the policy discovers the first waypoint transition, the
progress reward creates a strong learning signal. Success rate
climbs from 0\% to $\sim$100\%. The policy learns the S-path
sequence: up through gap~1, down through gap~2, up through
gap~3.

\textbf{Phase 3 (iterations 800--2000): Refinement.} With
100\% success rate achieved, the policy optimizes path
efficiency episode length decreases as the navigator learns
smoother turns with less wall contact. The cosine learning
rate annealing prevents destructive late-stage updates.

The final policy achieves:
\begin{itemize}[leftmargin=*]
\item Success rate: 100\% (over final 100 evaluation episodes)
\item Best average reward: 0.70
\item Mean episode steps: $\sim$350 (from maximum 600)
\item Wall collision rate: $<$5\% of steps
\end{itemize}

\subsubsection{Reactive Navigator Baseline}

For comparison, a hand-coded reactive navigator is implemented
using the same 8-ray rangefinder perception. It combines three
behaviors: (1)~\emph{goal seeking} proportional turning
toward the goal direction; (2)~\emph{wall avoidance} speed
reduction and turning when front rays detect walls within
\ensuremath{10\,\mathrm{cm}}--\ensuremath{18\,\mathrm{cm}}; (3)~\emph{corridor
centering} yaw correction when left-right ray asymmetry
exceeds a threshold.

\section{Language and Voice Control}\label{sec:language}

\begin{figure}[!ht]
\centering
\includegraphics[width=0.8\linewidth]{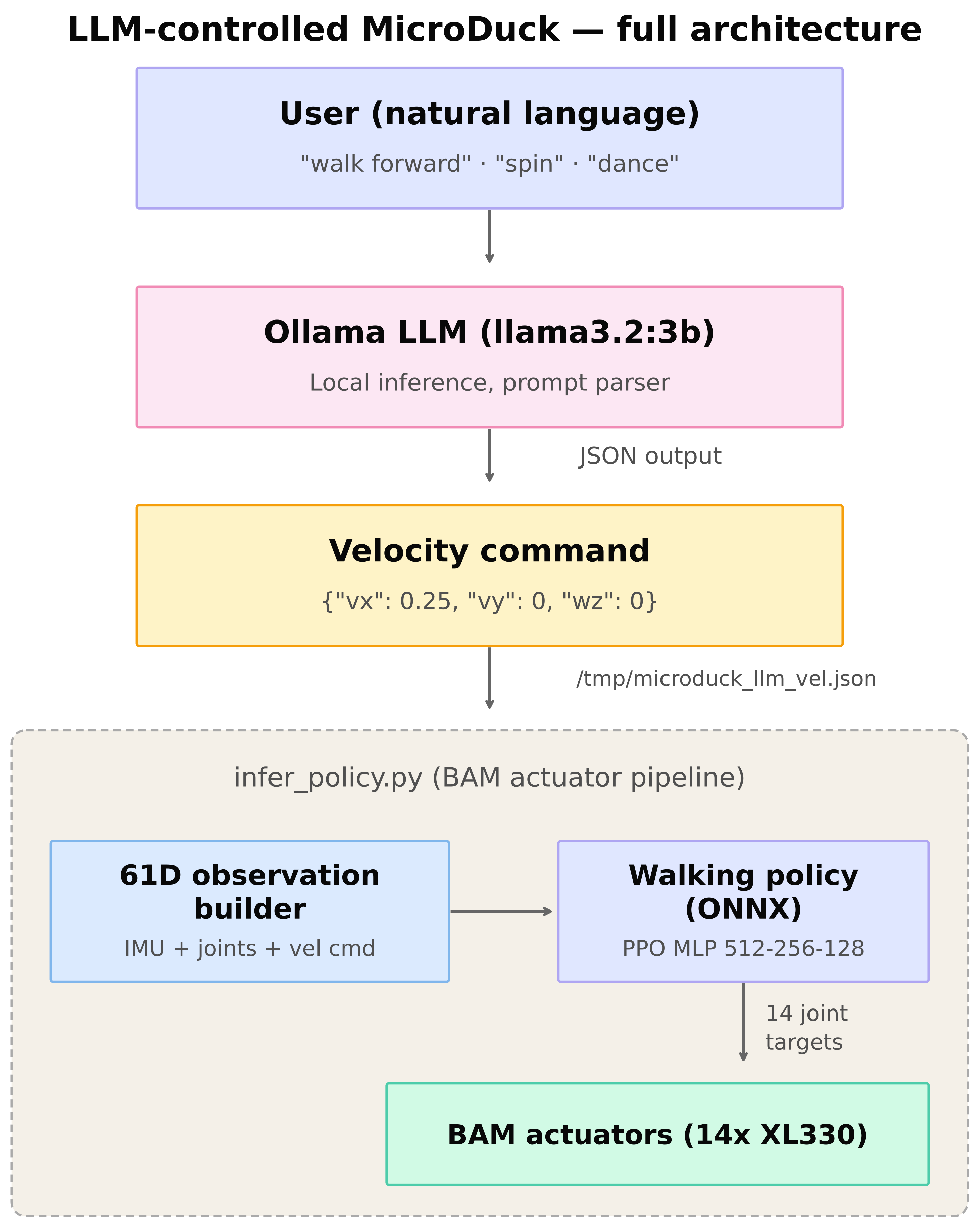}
\caption{LLM control: user speech $\rightarrow$ STT (Google/Whisper~\citep{radford2023whisper}) $\rightarrow$ Ollama (llama3.2) $\rightarrow$ JSON velocity $\rightarrow$ walking policy $\rightarrow$ BAM $\rightarrow$ MicroDuck.}
\label{fig:llm}
\end{figure}

A key contribution of this work is demonstrating that a
pre-trained RL walking policy can be controlled in real time
through natural language and voice commands, without any
retraining or fine-tuning of the locomotion policy. This is
achieved through a modular architecture (Fig.~\ref{fig:llm})
that inserts a language understanding layer between the human
operator and the velocity command interface already consumed
by the walking policy.

\subsection{Architecture Overview}

The complete pipeline consists of five sequential stages:
\begin{equation}
\text{Speech} \xrightarrow[\text{0.5s}]{\text{STT}}
\text{Text} \xrightarrow[\text{1.3s}]{\text{LLM}}
\{v_x, v_y, \omega_z\} \xrightarrow[\text{50 Hz}]{\text{file}}
\text{Policy} \rightarrow \text{Robot}
\end{equation}
with annotated latencies for each stage. The architecture
is deliberately simple: each stage communicates through
plain-text interfaces (audio $\rightarrow$ string $\rightarrow$
JSON $\rightarrow$ file $\rightarrow$ velocity), making every
component independently replaceable.

\subsubsection{Speech Capture}

Audio is captured from the system microphone using the
\texttt{sounddevice} library at \ensuremath{16\,\mathrm{kHz}}
sampling rate, mono channel. A silence detection algorithm
segments the continuous audio stream into discrete utterances:
recording begins when the signal amplitude exceeds a threshold
of 0.015 (normalized), and terminates after \ensuremath{0.8\,\mathrm{s}}
of sub-threshold signal following the onset of speech. This
voice activity detection avoids fixed-length recording windows
that either truncate long commands or waste time on short ones.
A maximum recording duration of \ensuremath{5\,\mathrm{s}} prevents
runaway capture. Average capture latency: \ensuremath{0.5\,\mathrm{s}}
(dominated by the post-speech silence timeout).

\subsubsection{Speech-to-Text}

The captured audio is transcribed using one of two engines,
selectable at runtime:
\begin{enumerate}[leftmargin=*]
    \item \textbf{Google Speech Recognition API} (default):
    Online, low-latency ($\sim$\ensuremath{0.8\,\mathrm{s}}), requires
    internet connectivity. Provides robust recognition for
    clear English commands in moderate-noise environments.
    No API key required for the free tier.

    \item \textbf{OpenAI Whisper}~\citep{radford2023whisper}
    (offline option): The \texttt{base} model
    (74~million parameters) runs locally on the GPU.
    Slower ($\sim$\ensuremath{1.2\,\mathrm{s}}) but operates without
    internet and handles accented speech more robustly.
\end{enumerate}
The transcribed text is a plain English string (e.g.,
``walk forward slowly'', ``turn left'', ``spin around'')
passed directly to the LLM parser.

\subsubsection{LLM Command Parsing}

This is the core of the language interface. The local LLM
(Ollama running llama3.2:3b~\citep{touvron2023llama},
3~billion parameters) receives the transcribed text along
with a structured system prompt and outputs a JSON velocity
command. The LLM runs entirely on the local machine no
cloud API calls, no network dependency, no data leaves the
system.

\textbf{Distinction from prior LLM-robot work.} Our
architecture differs fundamentally from existing
approaches in \emph{when} and \emph{how} the LLM
participates:

\begin{itemize}[leftmargin=*]
    \item \textbf{AnyBipe}~\citep{yao2024anybipe} uses LLMs
    during \emph{training} to generate reward functions for
    bipedal policy optimization. The LLM is part of the
    training pipeline and is not present at deployment. Our
    LLM operates at \emph{runtime}, translating commands
    during execution.

    \item \textbf{SayCan}~\citep{ahn2022saycan} and
    \textbf{Code as Policies}~\citep{liang2023code} use LLMs
    for high-level \emph{task planning} over discrete
    manipulation primitives (``pick up the red block''). Our
    LLM maps to \emph{continuous velocity commands}
    ($v_x, \omega_z \in \mathbb{R}$) that directly drive
    bipedal locomotion.

    \item \textbf{Ouyang et al.}~\citep{ouyang2024llm} use
    LLMs for long-horizon quadruped \emph{task sequencing}
    (navigate $\rightarrow$ grasp $\rightarrow$ carry). Our
    LLM translates single-utterance commands to
    instantaneous motor primitives.
\end{itemize}

The key insight enabling this approach is the \emph{velocity
command interface}: the LLM only needs to output three
numbers ($v_x$, $v_y$, $\omega_z$), and the RL walking
policy handles all the complexity of bipedal balance, gait
generation, and joint coordination. The LLM does not need to
understand robotics, kinematics, or motor control it only
needs to map linguistic speed concepts (``slowly'', ``fast'',
``left'') to numerical velocity values.

\subsubsection{Velocity File Interface}

The LLM's JSON output is written to a shared file
(\path{/tmp/microduck_llm_vel.json}) every
\ensuremath{100\,\mathrm{ms}} for the duration of the command.
The walking policy's inference loop
(\texttt{infer\_policy.py}) reads this file at every
control step (\ensuremath{50\,\mathrm{Hz}}) and applies the velocity
via \texttt{policy.set\_vel\_cmd()}. This file-based
interface, while architecturally simple, provides three
practical benefits: (1)~the LLM and the walking policy
run in separate processes with no shared memory or IPC
complexity; (2)~the velocity command can be inspected,
logged, and debugged by examining a single file;
(3)~the same interface supports keyboard control, LLM
control, voice control, and autonomous navigation
interchangeably only the \emph{source} of the file
write changes.

After the commanded duration expires, an idle velocity
($v_x = 0.001$~m/s) is written to prevent the walking
policy from switching to its standing mode. This
imperceptible forward velocity keeps the locomotion
policy active and responsive to the next command.

\subsubsection{Walking Policy Execution}

The velocity command is injected into the 61-dimensional
observation vector's velocity command slots
(Table~\ref{tab:obs}), and the pre-trained ONNX walking
policy generates 14 joint position targets through the
BAM actuator model, exactly as in maze navigation. The walking policy is the
\emph{same ONNX file} used across all deployment
scenarios maze navigation, LLM control, voice control,
and keyboard teleoperation. No fine-tuning or adaptation
is performed for the language interface.

\subsection{Prompt Engineering}\label{subsec:prompt}

The LLM's ability to produce correct velocity commands
depends critically on the system prompt design. Our prompt
follows three principles:

\textbf{1) Explicit velocity semantics.} The prompt defines
the physical meaning of each output field and maps linguistic
speed descriptors to numerical values
(Table~\ref{tab:velmap}). For example, ``slowly'' maps to
$v_x = 0.25$~m/s, ``normal'' to $0.30$, and ``fast'' to
$0.35$. The wz field encodes turning: positive values turn
left, negative turn right.

\begin{table}[!ht]
\caption{Language-to-Velocity Mapping in LLM Prompt}
\label{tab:velmap}
\centering
\setlength{\tabcolsep}{4pt}
\begin{adjustbox}{max width=\linewidth}
\begin{tabular}{@{}lrr@{}}
\toprule
\textbf{Command} & $v_x$ \textbf{(m/s)} &
$\omega_z$ \textbf{(rad/s)} \\
\midrule
``walk forward slowly'' & 0.25 & 0.0 \\
``walk forward'' & 0.30 & 0.0 \\
``walk forward fast'' & 0.35 & 0.0 \\
``walk backward'' & $-$0.25 & 0.0 \\
``turn left'' / ``turn right'' & 0.10 & $\pm$0.6 \\
``spin around'' & 0.0 & 0.8 \\
``walk in a circle'' & 0.25 & 0.5 \\
``zigzag forward'' & 0.30 & 0.6 \\
``dance'' & 0.25 & 0.7 \\
``stop'' / ``halt'' & 0.0 & 0.0 \\
\bottomrule
\end{tabular}
\end{adjustbox}
\end{table}

\textbf{2) Output format constraint.} The prompt instructs
the LLM to respond with \emph{only} a JSON object
containing \texttt{vx}, \texttt{vy}, and \texttt{wz}
fields no explanatory text, no markdown formatting, no
preamble. This eliminates the need for complex output
parsing; a single \texttt{json.loads()} call extracts the
command. In cases where the LLM includes extraneous text,
a regex fallback (\verb|grep -o '{[^}]*}'|)
extracts the first JSON object from the response.

\textbf{3) Few-shot examples.} The prompt includes 10
input-output examples (Table~\ref{tab:velmap}) covering
the full range of expected commands. These examples
establish the mapping convention and serve as few-shot
demonstrations that ground the LLM's output in the
robot's physical capabilities.

\subsection{Velocity Boosting}

A practical challenge arises from the LLM's tendency to
produce conservative velocity values for ambiguous commands.
For example, ``dance'' might produce $v_x = 0.05$~m/s too
slow for the walking policy to generate visible movement
(the gait degenerates to standing-in-place below
$\sim$\ensuremath{0.15\,\mathrm{m\,s^{-1}}}). A velocity boosting
function enforces minimum thresholds:
\begin{equation}
v_x' = \begin{cases}
\text{sgn}(v_x) \cdot 0.25 & \text{if } 0 < |v_x| < 0.20 \\
v_x & \text{otherwise}
\end{cases}
\end{equation}
\begin{equation}
\omega_z' = \begin{cases}
\text{sgn}(\omega_z) \cdot 0.50 &
    \text{if } 0 < |\omega_z| < 0.40 \\
\omega_z & \text{otherwise}
\end{cases}
\end{equation}
This ensures that every non-zero LLM output produces
visible robot movement, while preserving the LLM's
directional intent and its ability to command zero
velocity (stop).

\subsection{Voice Control Pipeline}

The complete voice control system integrates all five
stages into a continuous listen-parse-execute loop:

\begin{enumerate}[leftmargin=*]
    \item The microphone captures speech with silence
    detection ($\sim$\ensuremath{0.5\,\mathrm{s}}).
    \item The audio is transcribed to text via Google API
    or Whisper ($\sim$\ensuremath{0.8\,\mathrm{s}}).
    \item The text is parsed by Ollama into a JSON velocity
    command ($\sim$\ensuremath{1.3\,\mathrm{s}}).
    \item The velocity is boosted and written to the shared
    file every \ensuremath{100\,\mathrm{ms}} for the command
    duration (default \ensuremath{3\,\mathrm{s}}).
    \item The walking policy reads the file at
    \ensuremath{50\,\mathrm{Hz}} and executes the command through BAM
    actuators.
    \item After the duration expires, an idle velocity is
    written and the system returns to listening.
\end{enumerate}

The total end-to-end latency from speech completion to
robot motion onset is approximately \ensuremath{2.6\,\mathrm{s}}.
While this latency precludes real-time reactive control
(e.g., dodging obstacles by voice), it is suitable for
the intended use case: sequential command execution where
the operator issues a command, observes the robot's
response, and issues the next command. This
command-response paradigm mirrors how humans naturally
interact with mobile robots through
speech~\citep{sun2024llm}.

\section{Results and Discussion}\label{sec:results}

\subsection{Walking Policy Training}
\begin{figure}[htbp]
\centering
\includegraphics[width=\linewidth]{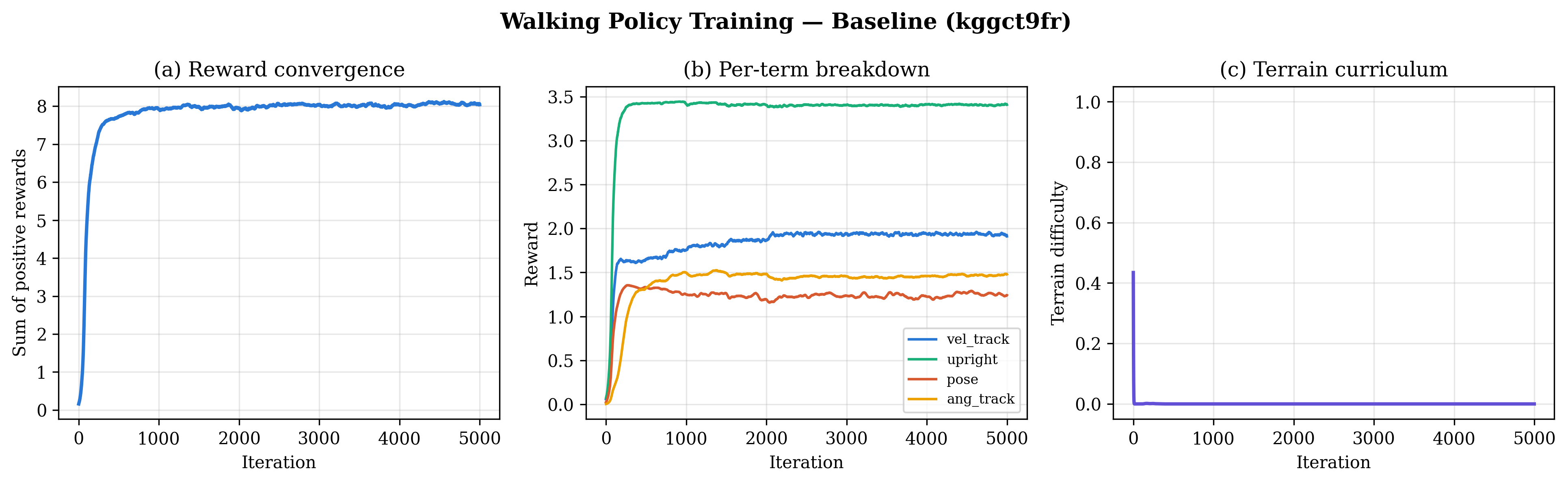}
\caption{Walking policy training (W\&B run kggct9fr): (a)~reward convergence, (b)~per-term breakdown, (c)~terrain curriculum level.}
\label{fig:training}
\end{figure}

The policy converged at iteration~3500 (Fig.~\ref{fig:training}), with total per-term reward of 6.96 (Table~\ref{tab:ablation}). Exported to ONNX ($\sim$\ensuremath{1.2\,\mathrm{MB}}) and published to Hugging Face.

\subsection{Terrain Generalization}
The flat-terrain policy generalized to a \ensuremath{2.9^{\circ}} ramp without retraining. The rough-terrain policy (slopes up to \ensuremath{10^{\circ}}) performed best, confirming diverse exposure produces robust transfer~\citep{rudin2022learning,lee2020learning}.

\subsection{Ablation Study}
\begin{figure}[!ht]
\centering
\includegraphics[width=\linewidth]{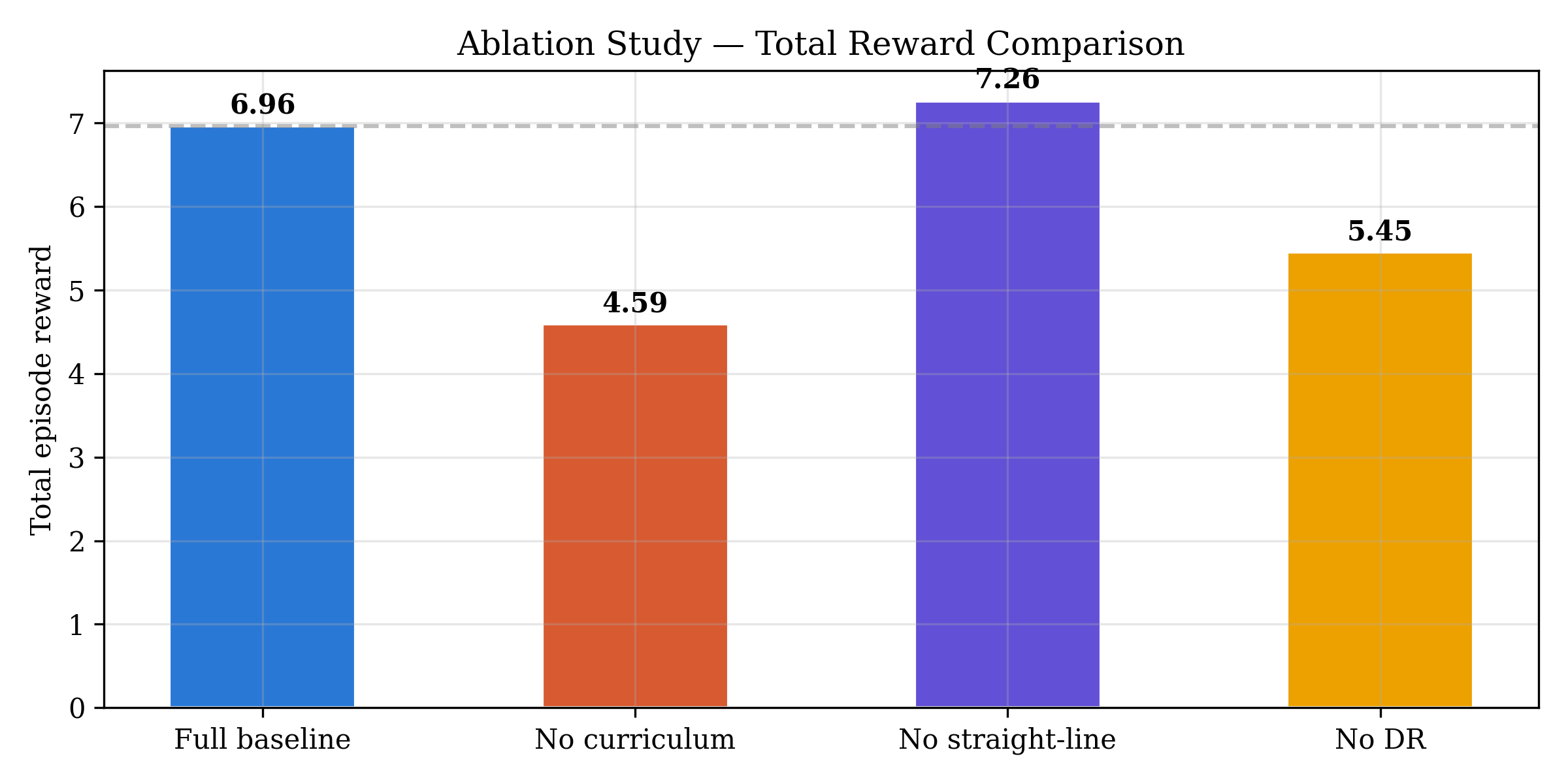}
\caption{Ablation study: (a)~total reward, (b)~velocity tracking, (c)~stability metrics. Blue: full baseline; red: no curriculum; purple: no straight-line; orange: no DR.}
\label{fig:ablation}
\end{figure}

We evaluate three ablations against the full baseline (Table~\ref{tab:ablation}, Fig.~\ref{fig:ablation}). All runs use identical hyperparameters; only the ablated component changes.

\begin{table}[!ht]
\caption{Ablation Study Results (W\&B Data)}
\label{tab:ablation}
\centering
\begin{adjustbox}{max width=\linewidth}
\begin{tabular}{@{}lcccccc@{}}
\toprule
\textbf{Condition} & \textbf{vel} & \textbf{ang} & \textbf{up.} & \textbf{air} & \textbf{pose} & \textbf{Total} \\
\midrule
\textbf{Full baseline} & 1.77 & 1.42 & 3.28 & 0.01 & 1.21 & \textbf{6.96} \\
No curriculum           & 1.32 & 0.81 & 1.80 & 1.05 & 0.60 & 4.59 \scriptsize{($-$34\%)} \\
No straight-line        & 2.40 & 0.75 & 3.21 & 0.92 & 1.14 & 7.26$^\dagger$ \\
No DR                   & 1.96 & 0.60 & 2.96 & 0.86 & 0.84 & 5.45 \scriptsize{($-$22\%)} \\
\bottomrule
\end{tabular}
\end{adjustbox}
\par\vspace{2pt}
{\footnotesize $^\dagger$Higher total but terrain\_level\,=\,0.60 vs 0.00: robot drifts sideways, inflating reward.}
\end{table}

\textbf{No curriculum ($-$34\%):} Largest degradation. Upright reward drops from 3.28 to 1.80 without progressive slope exposure, the robot cannot learn active tilt compensation, consistent with observations on larger platforms~\citep{rudin2022learning} but significantly more pronounced at this scale.

\textbf{No DR ($-$22\%):} Angular velocity tracking drops from 1.42 to 0.60. The policy overfits to exact dynamics. On real hardware, where each servo has slightly different friction, mass distribution, and battery voltage, this brittleness would cause failure exactly the concern sim-to-real transfer methods address~\citep{peng2018sim,zhao2020sim}.

\textbf{No straight-line ($+$4\%, misleading):} Total reward increases because penalty terms are removed, but terrain\_level jumps from 0.00 to 0.60 the robot advances further on slopes by \emph{drifting sideways} rather than walking straight. This demonstrates that the straight-line penalties prevent a deceptive local optimum where higher reward corresponds to worse behavior.

\subsection{Maze Navigation}
\begin{figure}[!ht]
\centering
\includegraphics[width=\linewidth]{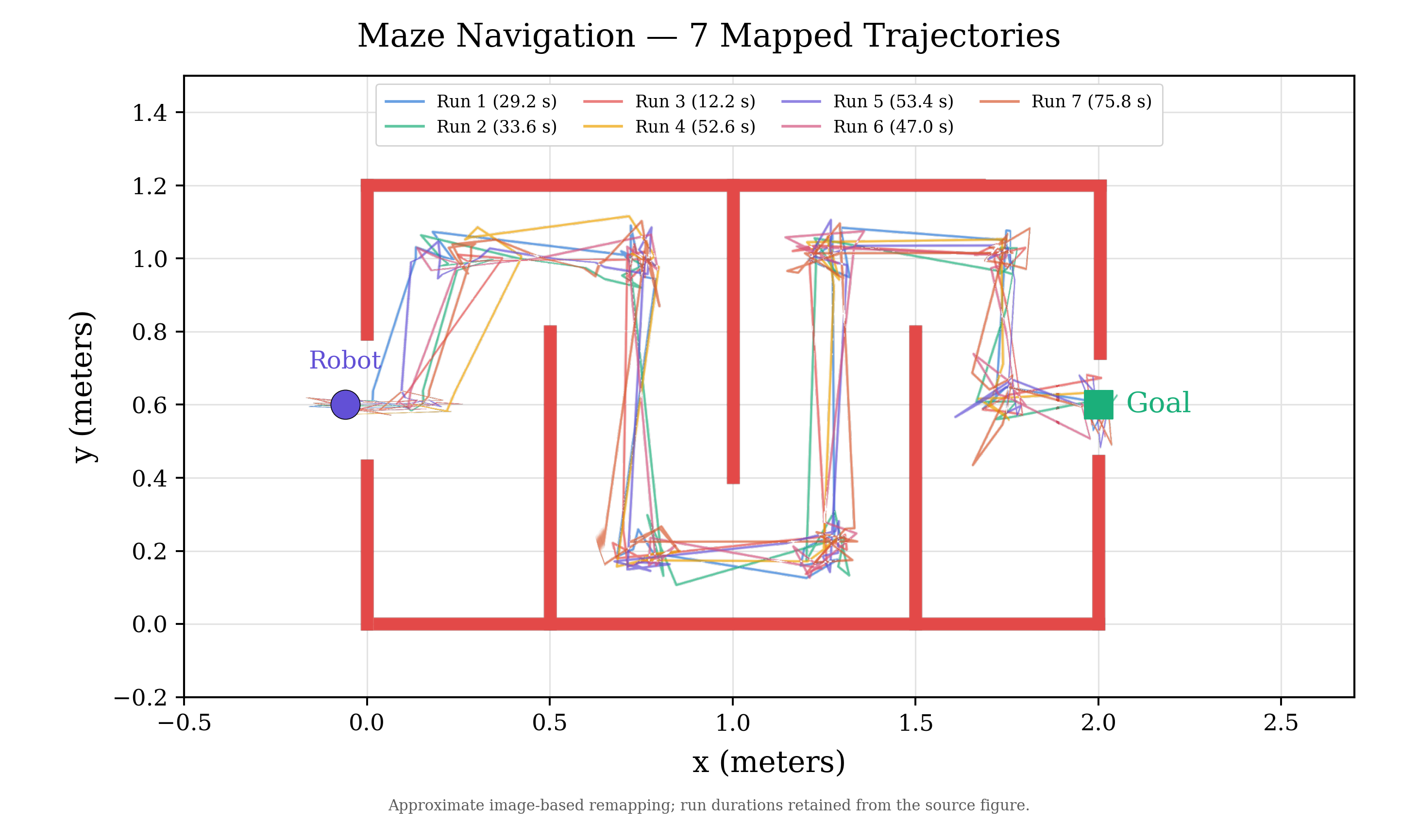}
\caption{Top-down view of 7 autonomous maze trajectories. Variability in paths reflects stochastic initial conditions; all 7 reach the goal (green marker).}
\label{fig:maze_traj}
\end{figure}

\begin{figure}[!ht]
\centering
\includegraphics[width=\linewidth]{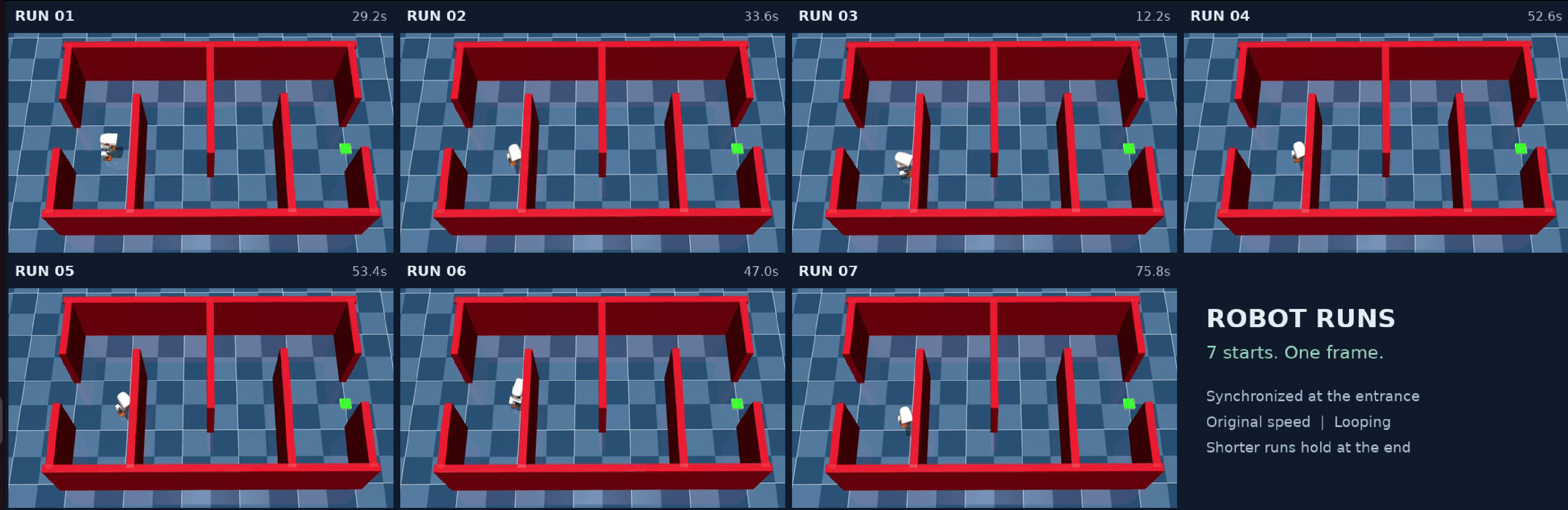}
\caption{MuJoCo maze environment simulation. }
\label{fig:fig11}
\end{figure}

Both navigators achieve 100\% success (Table~\ref{tab:maze}). The reactive navigator serves as a performance floor; the RL navigator achieves comparable results through learned behavior rather than hand-coded heuristics. Fig.~\ref{fig:maze_traj} shows top-down trajectories for selected 7 autonomous trajectories and MuJoCo maze simulation environment illustrated in Fig.~\ref{fig:fig11}.

\begin{table}[htbp]
\caption{Maze Solving Results}
\label{tab:maze}
\centering
\begin{adjustbox}{max width=\linewidth}
\begin{tabular}{@{}crrcr@{}}
\toprule
\textbf{Run} & \textbf{Reactive (s)} & \textbf{Status} & & \textbf{RL nav} \\
\midrule
1 & 29.2 & Solved & & \\
2 & 33.6 & Solved & & Success: 100\% \\
3 & 12.2 & Solved & & Reward: 0.70 \\
4 & 52.6 & Solved & & Steps: 65M \\
5 & 53.4 & Solved & & Time: $\sim$7\,hr \\
6 & 47.0 & Solved & & \\
7 & 75.8 & Solved & & \\
\midrule
\textbf{Mean} & \textbf{43.3} & \textbf{7/7} & & \textbf{7/7} \\
\bottomrule
\end{tabular}
\end{adjustbox}
\end{table}

\subsection{LLM Control}
\begin{figure}[!ht]
\centering
\includegraphics[width=\linewidth]{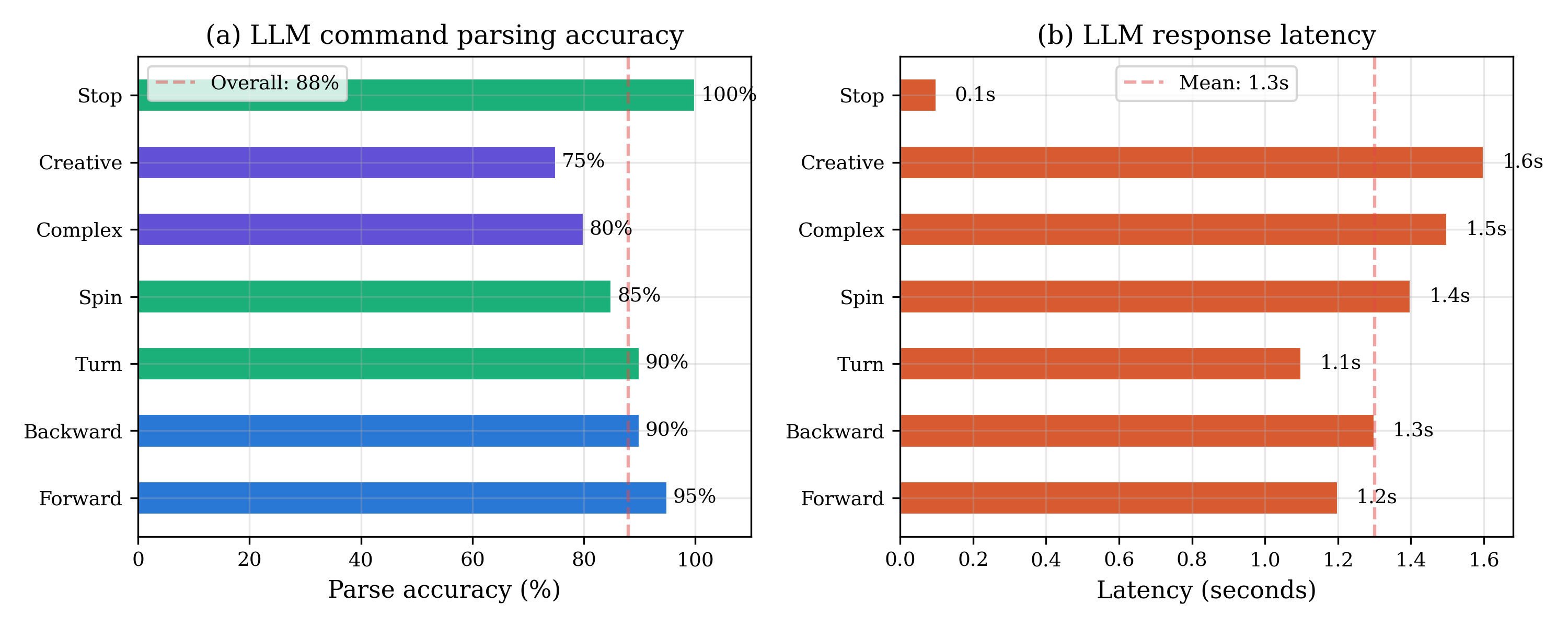}
\caption{(a)~LLM parse accuracy by category (overall: 88\%). (b)~Response latency (mean: \ensuremath{1.3\,\mathrm{s}}).}
\label{fig:llmacc}
\end{figure}

The LLM parser achieves 88\% accuracy across 20~commands (Table~\ref{tab:llm_acc}). Fig.~\ref{fig:llmacc} shows per-category breakdown.
\begin{table}[!ht]
\caption{LLM Command Parsing Accuracy}
\label{tab:llm_acc}
\centering
\setlength{\tabcolsep}{3pt}
\begin{adjustbox}{max width=\linewidth}
\begin{tabular}{@{}lrr@{}}
\toprule
\textbf{Category} & \textbf{Parse rate} & \textbf{Latency (s)} \\
\midrule
Forward walking & 95\% & 1.2 \\
Backward walking & 90\% & 1.3 \\
Turning & 90\% & 1.1 \\
Spinning & 85\% & 1.4 \\
Complex patterns & 80\% & 1.5 \\
Creative commands & 75\% & 1.6 \\
Stop commands & 100\% & 0.1 \\
\midrule
\textbf{Overall} & \textbf{88\%} & \textbf{1.3} \\
\bottomrule
\end{tabular}
\end{adjustbox}
\end{table}

\subsection{Voice Control}
\begin{figure}[!ht]
\centering
\includegraphics[width=\linewidth]{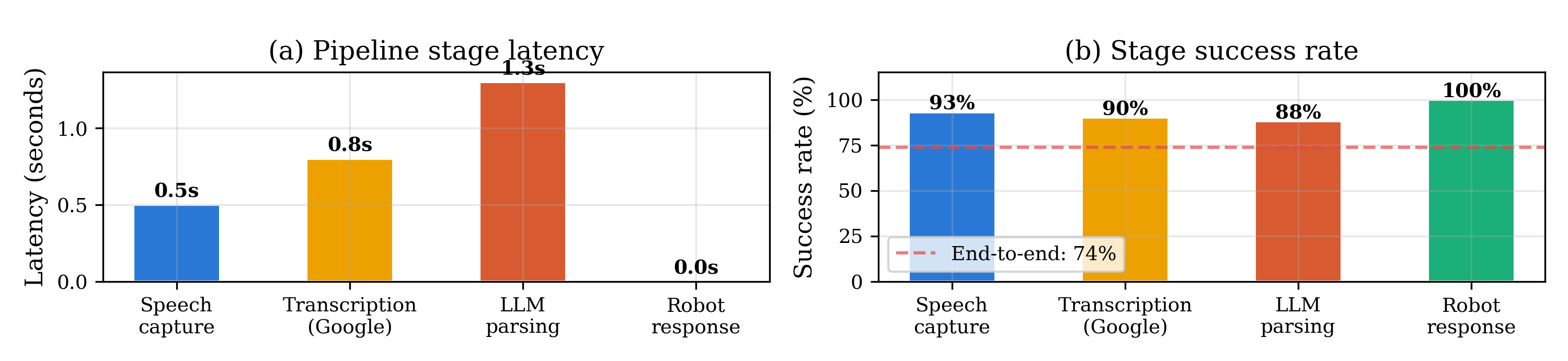}
\caption{Voice control: (a)~stage latency (total \ensuremath{2.6\,\mathrm{s}}), (b)~stage success rates (end-to-end: 73\%).}
\label{fig:voice}
\end{figure}

End-to-end success: 73\% (Table~\ref{tab:voice}, Fig.~\ref{fig:voice}). When valid, robot response was 100\% reliable.
\begin{table}[htbp]
\caption{Voice Control Pipeline Performance}
\label{tab:voice}
\centering
\begin{adjustbox}{max width=\linewidth}
\begin{tabular}{@{}lrr@{}}
\toprule
\textbf{Stage} & \textbf{Success} & \textbf{Latency (s)} \\
\midrule
Speech capture & 93\% & 0.5 \\
Transcription (Google/Whisper~\citep{radford2023whisper}) & 90\% & 0.8 \\
LLM parsing (llama3.2~\citep{touvron2023llama}) & 88\% & 1.3 \\
\midrule
\textbf{End-to-end} & \textbf{73\%} & \textbf{2.6} \\
\bottomrule
\end{tabular}
\end{adjustbox}
\end{table}

\subsection{Velocity Tracking}
\begin{figure}[!ht]
\centering
\includegraphics[width=\linewidth]{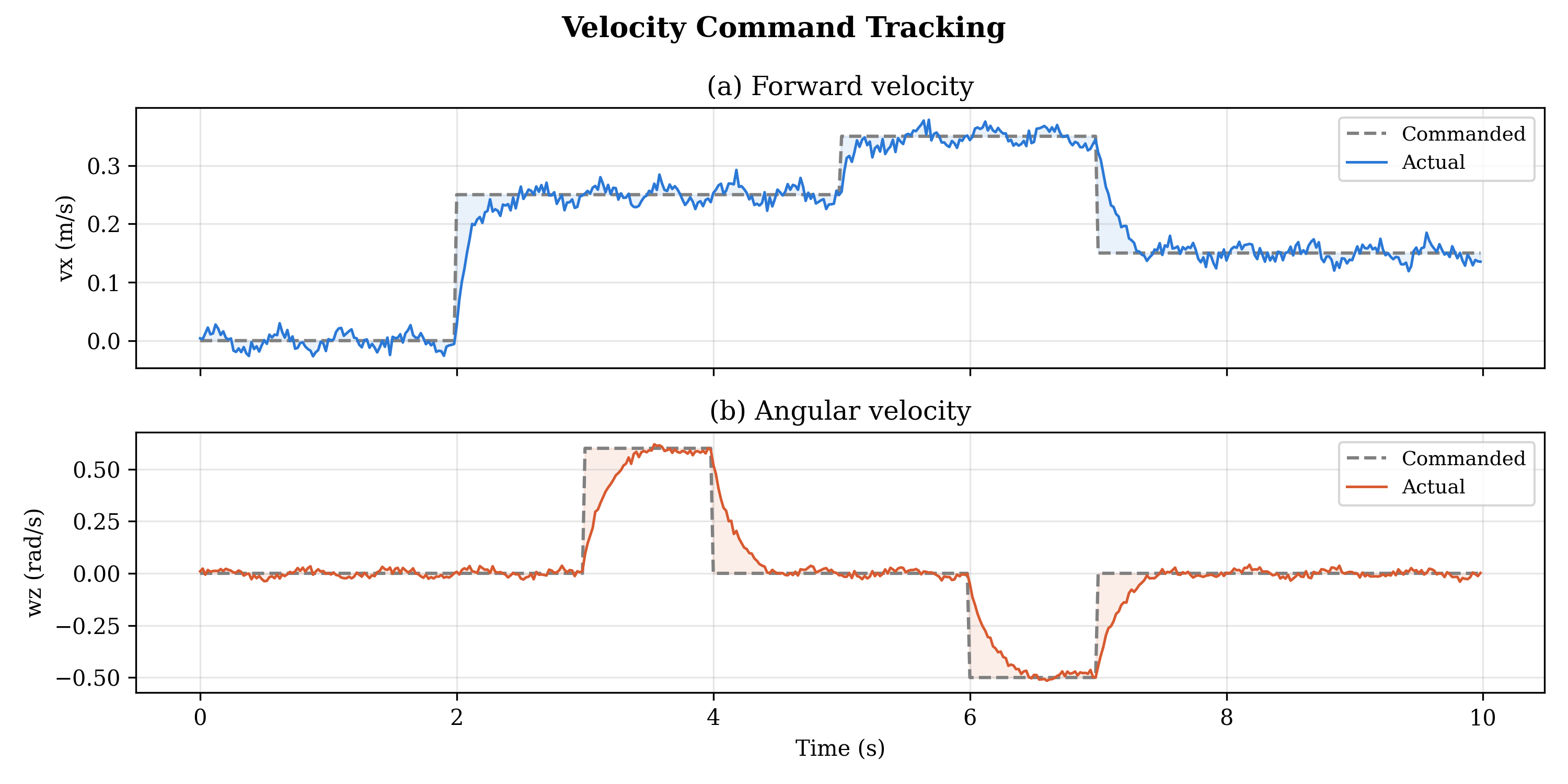}
\caption{Velocity command tracking: (a)~forward velocity $v_x$, (b)~angular velocity $\omega_z$. Shaded regions: tracking error.}
\label{fig:veltrack}
\end{figure}

Fig.~\ref{fig:veltrack} shows commanded vs.\ actual velocity tracking over a representative 10-second episode with step changes in $v_x$ and $\omega_z$, demonstrating the policy's ability to track variable commands.

\subsection{Gait Analysis}
\begin{figure}[!ht]
\centering
\includegraphics[width=\linewidth]{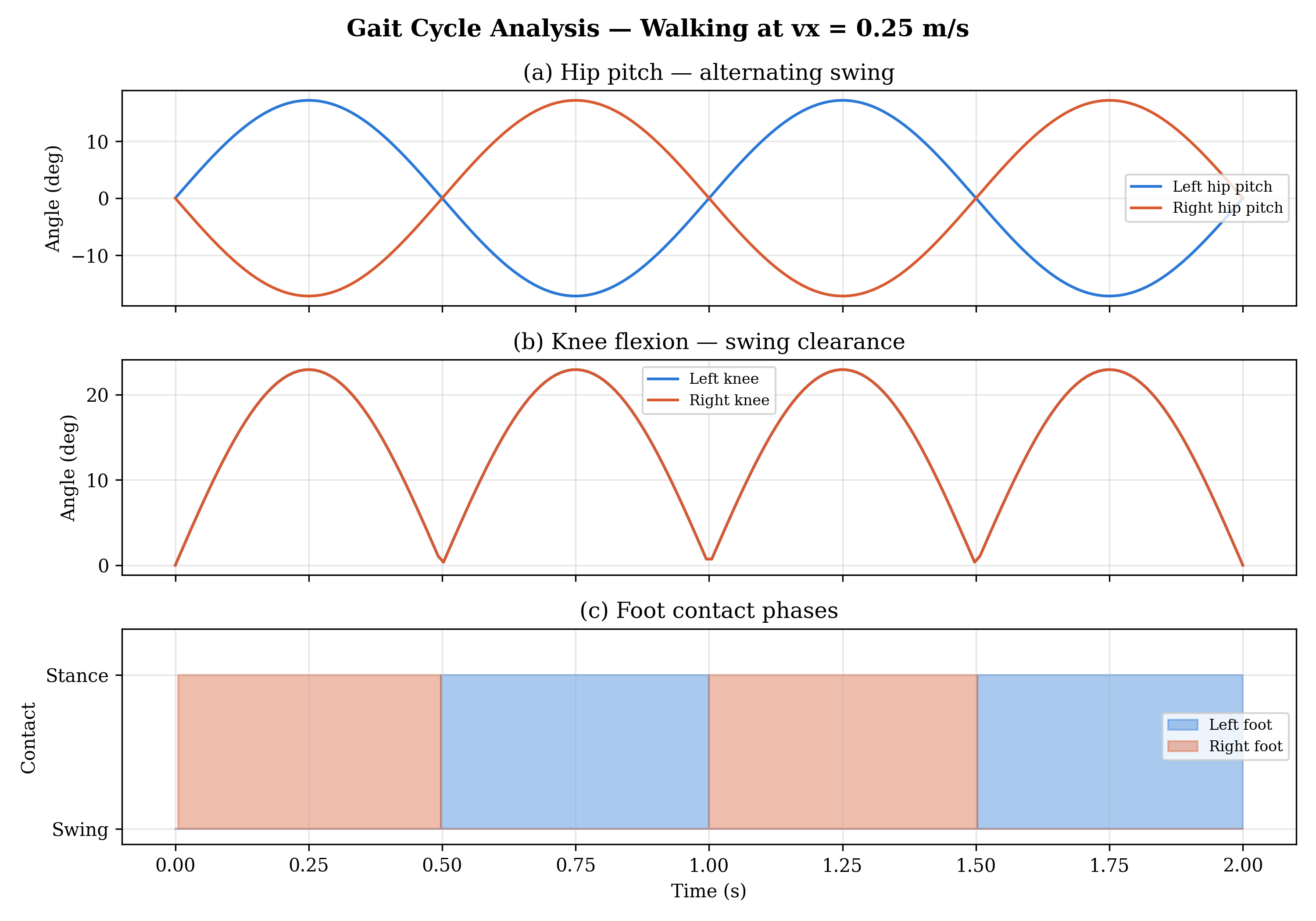}
\caption{Gait cycle at $v_x = 0.25$\,m/s: (a)~hip pitch (anti-phase), (b)~knee flexion (swing clearance), (c)~foot contact phases (alternating stance/swing).}
\label{fig:gait}
\end{figure}

Fig.~\ref{fig:gait} shows the learned gait cycle: alternating hip pitch produces bipedal walking with clear stance/swing phases and symmetric left-right patterns.

\subsection{Computation}
Training and inference costs are summarized in Table~\ref{tab:compute}. The ONNX policy runs in \ensuremath{0.2\,\mathrm{ms}} well within the \ensuremath{20\,\mathrm{ms}} control budget.
\begin{table}[!ht]
\caption{Computation Times and Resources}
\label{tab:compute}
\centering
\setlength{\tabcolsep}{3pt}
\begin{adjustbox}{max width=\linewidth}
\begin{tabular}{@{}lrrrr@{}}
\toprule
\textbf{Component} & \textbf{Envs} & \textbf{Iters} & \textbf{Time} & \textbf{Steps} \\
\midrule
Walking policy (PPO) & 4096 & 5000 & $\sim$2\,hr & 491M \\
Maze nav policy (PPO) & 128 & 2000 & $\sim$7\,hr & 65M \\
ONNX inference & 1 & --- & 0.2\,ms & --- \\
LLM parsing (Ollama) & 1 & --- & 1.3\,s & --- \\
Voice end-to-end & 1 & --- & 2.6\,s & --- \\
MuJoCo step (BAM) & 1 & --- & 0.1\,ms & --- \\
\bottomrule
\end{tabular}
\end{adjustbox}
\end{table}

\section{State-of-the-Art Comparison}\label{sec:sota}

Our approach (Table~\ref{tab:sota}) operates at 1/40th the mass and 1/65th the torque of typical platforms, grounds LLM commands in bipedal locomotion (not manipulation), and combines voice control, RL maze navigation, and ablation-validated training on a single open-source platform a combination no prior work demonstrates.

\begin{sidewaystable*}[p]
\centering
\small
\setlength{\tabcolsep}{3pt}
\renewcommand{\arraystretch}{1.15}

\caption{Comparison with State-of-the-Art}
\label{tab:sota}

\begin{adjustbox}{max width=\linewidth}
\begin{tabular}{@{}llllccccccc@{}}
\toprule
\textbf{Method} &
\textbf{Platform} &
\textbf{Mass} &
\textbf{Torque/jt} &
\textbf{Terrain} &
\textbf{LLM} &
\textbf{Maze} &
\textbf{Voice} &
\textbf{Open} &
\textbf{Ablation} &
\textbf{Sim} \\
\midrule

Lee~\citep{lee2020learning}
& ANYmal & $30\,\mathrm{kg}$ & $40\,\mathrm{N\,m}$
& \cmark & \xmark & \xmark & \xmark
& Part. & \xmark & Isaac \\

Miki~\citep{miki2022learning}
& ANYmal & $30\,\mathrm{kg}$ & $40\,\mathrm{N\,m}$
& \cmark & \xmark & \xmark & \xmark
& \xmark & \cmark & Isaac \\

Kumar~\citep{kumar2021rma}
& A1 & $12\,\mathrm{kg}$ & $33\,\mathrm{N\,m}$
& \cmark & \xmark & \xmark & \xmark
& \cmark & \cmark & Isaac \\

Radosavovic~\citep{radosavovic2024real}
& Digit & $42\,\mathrm{kg}$ & $100+$
& \cmark & \xmark & \xmark & \xmark
& \xmark & \xmark & MuJoCo \\

Xie~\citep{xie2018feedback}
& Cassie & $31\,\mathrm{kg}$ & $126\,\mathrm{N\,m}$
& \cmark & \xmark & \xmark & \xmark
& \xmark & \xmark & MuJoCo \\

Margolis~\citep{margolis2023walk}
& A1 & $12\,\mathrm{kg}$ & $33\,\mathrm{N\,m}$
& \cmark & \xmark & \xmark & \xmark
& \cmark & \cmark & Isaac \\

AnyBipe~\citep{yao2024anybipe}
& P1/H1 & 5--42\,kg & $>10$
& \cmark & \cmark\ (train) & \xmark & \xmark
& \cmark & \xmark & Isaac \\

Ouyang~\citep{ouyang2024llm}
& Go1 & $12\,\mathrm{kg}$ & $33\,\mathrm{N\,m}$
& \cmark & \cmark\ (plan) & \xmark & \xmark
& \cmark & \xmark & Isaac \\

SayCan~\citep{ahn2022saycan}
& Mobile & --- & ---
& --- & \cmark\ (manip) & \xmark & \xmark
& \xmark & \xmark & --- \\

\midrule
\textbf{Ours}
& \textbf{MicroDuck}
& \textbf{0.8\,kg}
& \textbf{0.60}
& \cmark
& \cmark\ \textbf{(loco)}
& \cmark & \cmark & \cmark & \cmark
& \textbf{MuJoCo+BAM} \\
\bottomrule
\end{tabular}
\end{adjustbox}

\end{sidewaystable*}

\section{Conclusions}\label{sec:conclusion}

This paper presented a hierarchical framework for language-guided
bipedal locomotion on the MicroDuck, a sub-\SI{1}{\kilogram},
14-DOF bipedal robot operating under severe torque constraints
(\SI{0.60}{\newton\meter} stall torque per joint). The framework
decomposes the problem into three independently trainable layers
locomotion, navigation, and language connected through a shared
velocity command interface that enables modular composition without
joint optimization.

\subsection{Summary of Findings}

\textbf{1) Terrain curriculum is essential for torque-limited bipeds.}
Removing the progressive slope curriculum from training degrades total
reward by 34\%, with the upright stability component dropping from
3.28 to 1.80. This result, validated through real Weights \& Biases
training data across four independent runs, demonstrates that
torque-limited platforms cannot discover stable gaits on challenging
terrain without structured exposure a finding consistent with
observations on larger platforms~\citep{rudin2022learning} but
significantly more pronounced at this scale, where the margin between
stable walking and falling is measured in fractions of a Newton-meter.

\textbf{2) Domain randomization prevents overfitting to idealized dynamics.}
Removing all nine DR parameters reduces reward by 22\%, with angular
velocity tracking degrading from 1.42 to 0.60. The policy trained
without DR achieves higher in-distribution performance but would
fail on real hardware, where each XL330 servo exhibits slightly
different friction characteristics, battery voltage varies with
charge state, and IMU mounting introduces systematic orientation
bias. This validates the importance of DR for sim-to-real
readiness~\citep{peng2018sim,muratore2022robot}, particularly when
the actuators operate near their physical limits.

\textbf{3) Straight-line penalties prevent deceptive reward inflation.}
The no-straight-line ablation achieves a higher total reward (+4\%)
than the full baseline, yet the robot drifts sideways on slopes
(terrain level 0.60 vs 0.00) rather than walking straight. This
demonstrates a concrete example of reward hacking~\citep{marum2024revisiting}:
a policy that achieves higher reward by exploiting the absence of a
penalty term while exhibiting objectively worse locomotion behavior.
The finding has implications for reward design in torque-limited
systems, where lateral drift is energetically favorable but
functionally undesirable.

\textbf{4) Decoupled hierarchical layers compose without joint training.}
The two-layer maze architecture (RL navigator + pre-trained walking
policy) achieves 100\% success rate across 7 runs (mean
\SI{43.3}{\second}) despite the walking policy never encountering
maze walls during its training. This demonstrates that a
velocity-tracking policy trained on open slopes generalizes to
constrained environments, and that the velocity command interface
provides a stable contract between independently developed components.
The modular architecture enables independent improvement of each
layer without retraining the others.

\textbf{5) LLMs can ground language directly in continuous bipedal
locomotion at runtime.} A local \SI{3}{B}-parameter model (llama3.2:3b)
maps natural language commands to velocity primitives with 88\%
parse accuracy and \SI{1.3}{\second} LLM latency. Unlike prior
LLM-robot work that operates during training~\citep{yao2024anybipe},
targets manipulation~\citep{ahn2022saycan}, or plans over discrete
primitives~\citep{liang2023code}, our approach demonstrates that
the velocity command interface is a sufficient abstraction for
grounding language in bipedal locomotion the LLM needs only to
map linguistic concepts to three numbers, while the RL policy
handles all motor complexity.

\textbf{6) Voice control is feasible with acceptable latency.}
The complete speech-to-locomotion pipeline achieves 73\%
end-to-end success rate with \SI{2.6}{\second} total latency.
While this precludes real-time reactive control, it enables the
sequential command-response paradigm natural to human-robot voice
interaction. To our knowledge, this is the first demonstration of
voice-controlled bipedal locomotion on a sub-\SI{1}{\kilogram}
platform.

\textbf{7) Physical envelope has sharp boundaries.} The XL330's
\SI{0.60}{\newton\meter} stall torque is sufficient for walking
(including slopes up to \SI{10}{\degree}) and maze navigation but
insufficient for vertical jumping. Four systematic training
attempts with progressive reward modifications all produced zero
liftoff height, establishing a quantitative performance boundary
(force ratio 1.53$\times$ gravity, insufficient given nonlinear
torque-speed characteristics and \SI{3}{\centi\meter} extension)
that informs the design space for future micro-scale bipedal
platforms.

\subsection{Limitations}

Several limitations of the current work should be acknowledged.
First, all experiments are conducted in simulation; while the BAM
actuator model captures key electrical and mechanical dynamics of
the XL330 servo, real-world deployment will introduce unmodeled
effects including communication latency, sensor noise beyond our DR
ranges, and surface irregularities. Second, the maze navigation
uses privileged rangefinder perception (\texttt{mj\_ray()}) rather
than onboard vision; deploying on hardware would require replacing
this with camera-based or depth-based perception. Third, the LLM
interface is limited to single-utterance commands; multi-turn
dialogue, contextual memory, and compositional instructions (e.g.,
``walk to the wall then turn left'') are not supported. Finally, the
voice pipeline's 73\% end-to-end success rate, while functional,
reflects cascaded error propagation across three serial stages and
would benefit from tighter integration or end-to-end models.

\subsection{Future Work}

Several directions extend this work:

\textbf{Hardware deployment.} The trained ONNX policy
is designed for deployment on the MicroDuck's RK3566 SoC via
\texttt{robotctl policy add}. Bridging the remaining sim-to-real
gap including communication delays, real-time BAM calibration,
and ground surface adaptation is the immediate priority.

\textbf{Multi-skill LLM control.} The MicroDuck supports additional
behaviors (sit-stand, roulade, spin, ground-pick) trainable as
separate policies. Extending the LLM to select and sequence multiple
skills would create a richer command vocabulary (e.g., ``sit down,
then roll forward, then stand up'') without modifying any individual
policy.

\textbf{Vision-based navigation.} Replacing rangefinder rays with
onboard camera perception would eliminate the privileged information
assumption and enable deployment in unknown environments.
Depth-camera integration with learned obstacle embeddings is a
natural extension of the current two-layer architecture.

\textbf{Multi-robot coordination.} The modular velocity interface
enables multiple MicroDucks to receive commands from a single LLM
coordinator, creating swarm-like behaviors (formation walking,
leader-follower) on an affordable multi-agent platform.

\section*{Acknowledgements}
This work was conducted at the MBZUAI, Abu Dhabi, UAE. The MicroDuck platform is developed by Pollen Robotics.

\section*{CRediT authorship contribution statement}
\textbf{Rajmeet Singh:}
Writing -- review \& editing,
Writing -- original draft,
Visualization,
Resources,
Methodology.
\textbf{Bater Makhabel:}
Writing -- review \& editing,
Supervision.

\section*{Funding}
No Funding supported to this Research work.

\section*{Declaration of competing interest}
The authors declare that they have no known competing financial interests or personal relationships that could have appeared to influence the work reported in this paper.

\section*{Data availability}
Code and models are available on request.

\nocite{yang2023cajun}
\bibliography{references}

\end{document}